\documentclass[12pt,journal,draftclsnofoot,onecolumn]{IEEEtran}
\usepackage[utf8]{inputenc}
\usepackage[T1]{fontenc}
\usepackage[numbers,square,comma,sort&compress]{natbib}
\usepackage{amsmath,amssymb,amsfonts,amsthm,mathtools}
\usepackage{algorithmic}
\usepackage{graphicx}
\usepackage{subfigure}
\usepackage{array,booktabs,arydshln}
\usepackage{capt-of}
\usepackage{float}
\usepackage{xcolor}
\usepackage{url}
\usepackage[letterpaper,margin=1in]{geometry}
\usepackage{indentfirst}
\usepackage{setspace}
\usepackage[hidelinks]{hyperref}
\usepackage{xr-hyper}
\graphicspath{{figures/}}
\floatstyle{ruled}
\newfloat{algorithm}{tbp}{loa}
\floatname{algorithm}{Algorithm}
\floatstyle{plain}

\def\BibTeX{{\rm B\kern-.05em{\sc i\kern-.025em b}\kern-.08em
    T\kern-.1667em\lower.7ex\hbox{E}\kern-.125emX}}
\newcommand{\mstd}[2]{#1{\normalfont\tiny(#2)}}
\newcommand{\algorithmicinput}{\textbf{Input:}}
\newcommand{\INPUT}{\item[\algorithmicinput]}

\theoremstyle{plain}
\newtheorem{theorem}{Theorem}[section]
\newtheorem{proposition}[theorem]{Proposition}

\newtheorem{corollary}[theorem]{Corollary}

\theoremstyle{definition}
\newtheorem{remark}[theorem]{Remark}
\newtheorem{assumption}[theorem]{Assumption}
    
\begin{document}

\title{Generative Flows for Potential Outcomes and Counterfactuals}

\author{Dongze Wu, David I. Inouye, Yao Xie%
\thanks{D. Wu and Y. Xie are with the School of Industrial and Systems Engineering,
Georgia Institute of Technology, Atlanta, GA, USA (e-mail: dwu381@gatech.edu;
yao.xie@isye.gatech.edu). D. Inouye is with the Elmore Family School of Electrical and Computer Engineering, Purdue University, West Lafayette, IN, USA (e-mail: dinouye@purdue.edu).}%
}

\maketitle

\begin{abstract}
Average treatment effects summarize how interventions change outcomes, but individualized decisions require predictions and outcome distributions of what would have happened to the same individual under an alternative treatment. We develop PO-Flow, a continuous normalizing flow framework that represents treatment-specific outcome distributions through invertible transport maps. It supports potential-outcome generation, conditional average treatment-effect estimation, factual-conditioned counterfactual prediction, and density evaluation. To address treatment imbalance in observational data, we incorporate cross-fitted inverse-probability weighting into the flow-matching loss. We establish sufficient conditions under which encoding a factual outcome and decoding under an alternative treatment recovers the individual counterfactual. Under shared exogenous noise and a common encoded reference distribution, scalar recovery follows from increasing noise-to-latent maps. For multivariate outcomes, we assume triangularity or a convex-gradient structure. These conditions distinguish fitting outcome distributions from recovering their causal correspondence. Experiments on four scalar benchmarks and seven multivariate or image settings, with dimensions up to 784, compare against ten scalar and five multivariate baselines. Controlled ablations show IPW improves all reported scalar counterfactual RMSE, PEHE, and potential-outcome RMSE. PO-Flow achieves the lowest PEHE and potential-outcome RMSE on all four scalar benchmarks and attains or ties the lowest mean in 26 of 35 multivariate dataset--metric comparisons.
\end{abstract}

\begin{IEEEkeywords}
Causal inference, potential outcomes, counterfactual prediction, flow generative models,
inverse probability weighting, transport maps.
\end{IEEEkeywords}

\section{Introduction}

Causal inference studies how interventions change outcomes. A central task is treatment-effect estimation: understanding how an outcome differs under treatment and control \cite{rubin1974estimating,holland1986statistics}. Much of the literature estimates average effects, either over a population or conditional on observed covariates \cite{imbens2004nonparametric,athey2016recursive,kunzel2019metalearners}. These quantities are useful for evaluating treatments, but individualized decisions often require more information. They depend on how outcomes vary across individuals, the risk of adverse responses, and what an alternative treatment would mean for a particular person \cite{feuerriegel2024causal}. This motivates moving beyond average effects toward learning outcome distributions and predicting individual counterfactual outcomes.

\begin{figure}[tbp]
\centering
\subfigure[Average treatment-effect estimation.\label{fig:average-effects}]{%
  \includegraphics[width=0.35\linewidth]{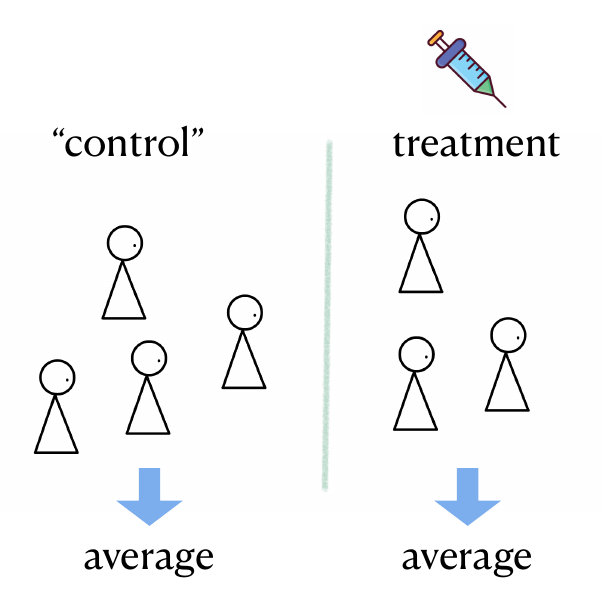}}
\subfigure[Individual counterfactual prediction.\label{fig:individual-counterfactual}]{%
  \includegraphics[width=0.35\linewidth]{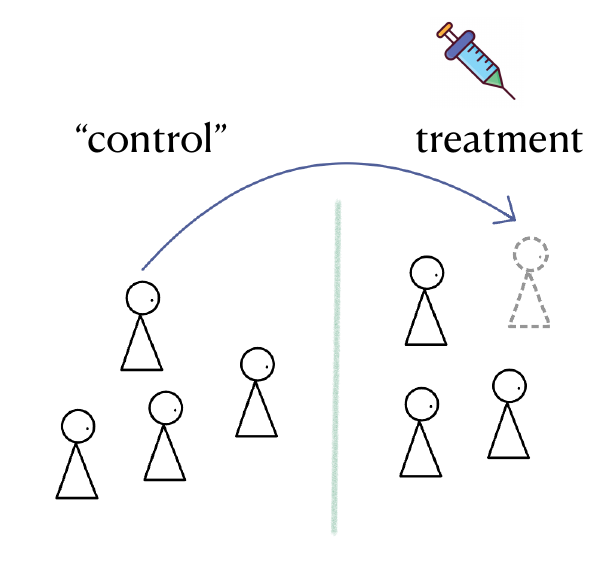}}
\caption{From average effects to individual counterfactuals. (a) ATE and CATE summarize treatment effects on average, overall and for people with the same covariates. (b) Given a person's observed outcome under one treatment, we ask what would have happened to that same person under the other. The dashed person marks the unobserved outcome; the arrow shows a pairing. Outcome distributions describe variation and tail risk, but identifying this pairing requires additional structural assumptions.}
\label{fig:average-to-counterfactual}
\end{figure}

In the potential outcomes framework, an individual with pretreatment covariates $X$ has two outcomes: $Y^{(1)}$ under treatment and $Y^{(0)}$ under control. Only one is observed \cite{holland1986statistics}. The average treatment effect (ATE), $\mathbb{E}[Y^{(1)}-Y^{(0)}]$, summarizes the population-level difference, while the conditional average treatment effect (CATE), $\mathbb{E}[Y^{(1)}-Y^{(0)}\mid X]$, describes how the mean difference varies with covariates. Even a CATE estimate remains an average among individuals with the same observed characteristics. Figure~\ref{fig:average-to-counterfactual}(a) illustrates this focus on average effects. These summaries do not describe the full range of possible outcomes under either treatment. Two treatments may have similar expected outcomes but different variability or tail risks. Learning the treatment-specific distributions $p(Y^{(a)}\mid X)$ provides this additional information \cite{zhang2013assessing,johansson2022generalization}.

An individual counterfactual asks a further question. After observing a person's outcome under one treatment, what would have happened to that same person under another? For example, a patient's observed response to one therapy may provide information relevant to predicting their response to an alternative. We refer to this as \emph{factual-conditioned counterfactual prediction}, with target $Y^{(1-a)}\mid X=x,Y^{(a)}=y$.\footnote{Some works \cite{johansson2016learning,lei2021conformal} use ``counterfactual'' for the unobserved potential outcome predicted from covariates alone, without conditioning on the factual outcome.} The modeling challenge is that the two outcome distributions do not specify how the responses are related for the same individual. Under consistency, unconfoundedness, and overlap, observational data identify the conditional distribution under each treatment, but generally do not identify the joint distribution of the two potential outcomes \cite{nasr2023counterfactual,de2024transport}. Therefore, factual-conditioned prediction requires specifying a correspondence across treatments and understanding the assumptions that justify it. Figure~\ref{fig:average-to-counterfactual}(b) illustrates the individual question: the observed response is linked to an unobserved response under the alternative treatment. The arrow represents the correspondence to be modeled, whose causal interpretation requires additional structural assumptions.

Transport maps provide a way to represent this correspondence. For fixed covariates, a map from the control outcome distribution to the treated outcome distribution specifies how an outcome under control is matched to an outcome under treatment. The resulting pairing defines a joint model of the two potential outcomes. Optimal transport chooses such a pairing by minimizing a prescribed cost \cite{charpentier2023optimal,de2024transport}. Balakrishnan, Kennedy, and Wasserman use this viewpoint to construct conservative counterfactual laws compatible with the observed distributions \cite{balakrishnan2025conservative}. From this perspective, counterfactual modeling involves both learning the outcome distributions and specifying how they are connected. The connection is a modeling choice: a map can match the observed distributions without recovering the actual pairing of outcomes for each individual.

Flow-based generative models make these maps practical to learn. A continuous normalizing flow (CNF) represents an invertible transformation through a neural ordinary differential equation \cite{chen2018neural}. Conditioned on covariates and treatment, the flow maps a simple reference distribution to the corresponding outcome distribution. Let $G_{x,a}$ denote this map under treatment $a$. Sampling $Z\sim q$ and evaluating $G_{x,a}(Z)$ generates a potential outcome. Given an observed outcome $y$, the inverse map gives a reference value $G_{x,a}^{-1}(y)$, which can then be mapped to the outcome under the alternative treatment:
\[
T_{x,a\to 1-a}(y)
=G_{x,1-a}\!\left(G_{x,a}^{-1}(y)\right).
\]
This encode--decode map specifies a candidate correspondence across treatments and gives counterfactual predictions. Flow matching learns the underlying velocity field by regression, without simulating the learned ODE during training \cite{lipman2023flow,albergo2022building}.

Learning these maps from observational data brings two distinct challenges. First, treatment assignment changes which outcomes are observed across the covariate space. Some treatment--covariate regions contribute relatively few observations to an unweighted training loss, making the allocation of fitting error important in finite samples. Generative approaches address observational bias through interventional density corrections, orthogonal diffusion losses, or inverse-probability-weighted objectives for longitudinal outcomes \cite{melnychuk2023normalizing,ma2024diffpo,wu2024counterfactual}. We incorporate cross-fitted inverse-probability weighting (IPW) directly into conditional flow matching to rebalance training across treatments in the static potential outcomes setting.

Second, fitting the outcome distributions does not determine whether factual encoding recovers the same individual information across treatments. We study sufficient conditions under which the encode--decode map preserves the individual's exogenous noise and recovers the counterfactual. These conditions apply to the noise-to-latent maps and are separate from the weighted training objective. We develop both components within PO-Flow: observational weighting for distribution learning, and a structural analysis of when the learned correspondence supports individual counterfactual recovery.

\subsection{Main Contributions}

We develop PO-Flow for learning potential-outcome distributions and predicting factual-conditioned counterfactual outcomes from observational data. A single invertible model supports potential-outcome generation, CATE estimation, counterfactual prediction, and density evaluation. Our main contributions are:
\begin{itemize}
\item[(i)] We incorporate cross-fitted inverse-probability weighting into the flow-matching loss to account for treatment imbalance in observational data. The weighted objective rebalances the contribution of treatment--covariate regions to training. In controlled ablations, IPW improves all reported scalar counterfactual RMSE, PEHE, and potential-outcome RMSE results.
\item[(ii)] We establish sufficient conditions for exact counterfactual recovery for scalar and multivariate outcomes. Under shared exogenous noise and a common encoded reference distribution, scalar recovery follows when the noise-to-latent maps are increasing. For multivariate outcomes, we give two alternatives: lower-triangular maps with conditional coordinatewise monotonicity in a common order, or maps that are gradients of convex functions. These results clarify when matching outcome distributions also recovers the individual correspondence across treatments.
\item[(iii)] We evaluate PO-Flow on four scalar benchmarks and seven multivariate or image settings, with outcome dimensions up to 784, comparing against ten scalar and five multivariate baselines. The evaluation covers potential-outcome prediction, CATE estimation, factual-conditioned counterfactual prediction, and distribution modeling. PO-Flow achieves the lowest PEHE and potential-outcome RMSE on all four scalar benchmarks, and attains or ties the lowest mean in 26 of the 35 reported multivariate dataset--metric comparisons.
\end{itemize}

The earlier conference version \cite{wu2026flow} introduced the PO-Flow framework and its encode--decode construction, together with a supporting scalar recovery guarantee. This journal version adds cross-fitted inverse-probability-weighted training, extends the recovery analysis to multivariate outcomes, and expands the empirical evaluation with multivariate and image settings and controlled IPW ablations.

\subsection{Literature Review}

\textit{Treatment-Effect Estimation:}
Methods for estimating heterogeneous treatment effects include meta-learners (S, T, X, and DR-Learner), which fit outcome regressions or regress pseudo-outcomes on covariates \cite{kunzel2019metalearners,curth2021nonparametric,kennedy2023towards}. Bayesian causal forests separate treatment-effect heterogeneity from prognostic variation \cite{hahn2020bayesian}. Neural methods include TARNet, which fits treatment-specific outcome regressions, and CFR, which also balances representations across treatment groups \cite{shalit2017estimating}. DragonNet jointly estimates outcomes and propensity scores, with optional targeted regularization \cite{shi2019adapting}. These methods mainly estimate conditional means. PO-Flow models outcome distributions and estimates their means by sampling.

\textit{Potential-Outcome Prediction and Uncertainty:}
Predicting potential outcomes and estimating treatment effects are related but distinct tasks \cite{johansson2022generalization}. GANITE uses the factual outcome to impute counterfactuals adversarially, then uses an inference network to estimate individual effects from covariates \cite{yoon2018ganite}. CEVAE models latent confounders and observed proxies with a variational generative model \cite{louizos2017causal}. In our setting, treatment confounders are observed. Some work constructs conformal intervals for potential outcomes and individual effects \cite{lei2021conformal}. These intervals quantify predictive uncertainty but do not identify the joint distribution of potential outcomes. PO-Flow learns treatment-specific distributions and examines when an invertible correspondence between them can recover counterfactuals given a factual outcome.

\textit{Longitudinal Counterfactual Prediction:}
The Counterfactual Recurrent Network and Causal Transformer use longitudinal covariates, treatments, and outcomes to predict outcomes under alternative treatment sequences \cite{bica2020estimating,melnychuk2022causal}. Wu et al. combine a conditional generative model with inverse-probability reweighting to address the mismatch between observed and counterfactual outcome distributions under time-varying treatments \cite{wu2024counterfactual}. These methods study sequential treatment settings. Our query instead concerns a static binary treatment and conditions on pretreatment covariates and the realized factual outcome.

\textit{Structural Causal Models and Generative Counterfactuals:}
Deep structural causal models, causal normalizing flows, and neural causal models rely on structural assumptions for counterfactual inference \cite{pawlowski2020deep,javaloy2023causal,xia2023neural,nasr2023counterfactual}. Counterfactuals can be identified in bijective causal models under specified structures, including settings with unobserved confounding \cite{nasr2023counterfactual}. Causal flows use invertible mechanisms for interventional and counterfactual queries \cite{javaloy2023causal}. Diffusion causal models recover the factual latent variable through a deterministic forward process, then intervene during the reverse process \cite{sanchez2022diffusion}. We adapt latent factual encoding to treatment-specific potential-outcome distributions and add observational treatment weighting to flow matching. Our theoretical recovery conditions concern the noise-to-latent maps. Matching distributions alone does not identify which potential outcomes belong to the same individual.

\textit{Transport-Based Counterfactuals:}
Knowing the outcome distribution under each treatment does not tell us which two outcomes belong to the same person. Transport offers a way to specify that pairing. Some work uses optimal transport to match covariate distributions under different interventions \cite{charpentier2023optimal}. In our setting, pretreatment covariates stay fixed, and we need to pair potential outcomes instead. Other work examines how transport-based counterfactuals relate to structural causal counterfactuals \cite{de2024transport}. When the joint distribution is not identified, optimal transport can also provide conservative laws \cite{balakrishnan2025conservative}. For multivariate outcomes, continuous-time flows and dynamic optimal transport have been used to study identification and learn counterfactual maps by flow matching \cite{ribeiro2025dynamic}. PO-Flow estimates treatment-specific outcome distributions with cross-fitted IPW. Its sufficient recovery conditions rely on unique scalar increasing, triangular, or convex-gradient noise-to-latent maps. The training objective does not impose these optimal-transport coupling.

\textit{Flow and Diffusion Models for Outcome Distributions:}
Interventional Normalizing Flows estimate interventional outcome densities using nuisance and target flows with a one-step correction for doubly robust estimation \cite{melnychuk2023normalizing}. DiffPO learns covariate-conditioned potential-outcome distributions through a conditional diffusion model and an orthogonal diffusion loss that addresses treatment-selection bias \cite{ma2024diffpo}. Both therefore address observational bias, using different objectives from PO-Flow. Inverse-probability weighting for generative outcome modeling has also been studied in the longitudinal setting \cite{wu2024counterfactual}. We incorporate cross-fitted inverse-probability weights directly into conditional flow matching for static treatment-specific outcome distributions, building on established propensity-weighting methods \cite{hirano2003efficient,cole2008constructing}. Alongside outcome generation and density evaluation, the fitted flow supplies an encode--decode map for factual-conditioned prediction, with a separate structural analysis of its recovery properties.

The rest of the paper is organized as follows.
Section~II reviews the preliminaries.
Section~III formulates the potential-outcome and factual-conditioned
counterfactual prediction targets and states the standard causal
identification assumptions.
Section~IV presents PO-Flow, including inverse-probability-weighted
training and the encode--decode construction.
Section~V establishes sufficient conditions for scalar and multivariate
counterfactual recovery and derives the potential-outcome density formula.
Section~VI describes the benchmarks, evaluation metrics, and
experimental results.
Section~VII discusses the limitations and scope of the method and concludes the paper.

\section{Preliminaries}

We use continuous normalizing flows to move between two representations of an outcome: its observed value and a reference variable that is easy to sample. The ODE defines the invertible map, while flow matching provides a way to learn it. We review these two components before conditioning them on covariates and treatment. Starting from an outcome $y(0)=y_0 \in \mathbb{R}^{d_Y}$ at time $t=0$, the trajectory $y(t)$
evolves according to
\begin{equation}
\frac{dy(t)}{dt} = v_{\theta}(y(t), t), \quad t \in [0,1],
\label{ODE equation}
\end{equation}
where $v_{\theta}:\mathbb{R}^{d_Y} \times [0,1] \to \mathbb{R}^{d_Y}$ denotes the velocity
field parameterized by a neural network with parameters $\theta$.

Under standard regularity conditions, the Neural ODE induces an invertible mapping from
the data $y_0$ at $t=0$ to a base variable $y_1=z$ at $t=1$. The inverse mapping from $y_1$
to $y_0$ is obtained by integrating \eqref{ODE equation} backward in time. Throughout
this paper, we adopt the standard choice $y_1=z\sim\mathcal N(0,I_{d_Y})$ at $t=1$ and
denote the corresponding base density by $q$.

We train the CNF using \emph{conditional flow matching} (CFM)
\cite{lipman2023flow,albergo2022building}, which learns the velocity field without simulating the learned ODE during training.
Given a data sample $y_0$ and an independent base sample $y_1 \sim q(\cdot)$,
CFM defines a reference path $\phi_t = \phi(y_0, y_1; t)$ connecting $y_0$ and $y_1$.
We adopt the linear interpolant
\begin{equation}
\phi_t = (1-t)\,y_0 + t\,y_1,
\label{linear interpolant}
\end{equation}
This path moves at constant velocity from the observed outcome to the reference draw, so its associated reference velocity is
\begin{equation}
\frac{d\phi_t}{dt} = y_1 - y_0.
\label{ref v}
\end{equation}
A randomly selected point on this path supplies a training input, and the known path velocity supplies its regression target. CFM trains $v_\theta$ by minimizing
\begin{equation}
\mathcal{L}_{\mathrm{CFM}}(\theta) 
=
\mathbb{E}_{t \sim \mathcal{U}[0,1],\, y_0 \sim p_{\text{data}},\, y_1 \sim q(\cdot)}
\left\| v_\theta(\phi_t, t) - \frac{d\phi_t}{dt} \right\|^2.
\label{original FM loss}
\end{equation}
The reference paths provide training pairs; sampling follows the learned velocity field rather than any one of these paths. After training, drawing $y_1\sim q$ and integrating~\eqref{ODE equation} backward produces an outcome sample. Forward integration encodes an observed outcome into the same reference space.

\section{Formulation}
\label{sec:problem-formulation}

We consider observational data in which treatment assignment depends on pretreatment covariates. Each individual has covariates $X\in\mathbb R^{d_X}$, a binary treatment $A\in\{0,1\}$, and potential outcomes $Y^{(0)},Y^{(1)}\in\mathbb R^{d_Y}$. We observe an i.i.d. dataset $
\mathcal D=\{(y_i,x_i,a_i)\}_{i=1}^n,\ y_i=y_i^{(a_i)}.$
The missing outcome creates two related modeling questions: how to learn the outcome distribution under each treatment, and how to predict an individual's alternative outcome after observing the factual response. These questions require different information and have different identification requirements.

\subsection{Potential-Outcome Distributions}

Our first target is the treatment-specific conditional law
\[
p(Y^{(a)}\mid X=x),\qquad a\in\{0,1\}.
\]
For example, before a patient receives treatment, this distribution describes the range of possible responses among patients with the same observed characteristics. It supports sampling, uncertainty assessment, and comparison of treatments beyond their mean outcomes. Its conditional means also determine the CATE,
\[
\tau(x)=\mathbb E[Y^{(1)}\mid X=x]
       -\mathbb E[Y^{(0)}\mid X=x].
\]
Thus, learning both outcome distributions also provides a route to conditional mean treatment-effect estimation.

\subsection{Factual-Conditioned Counterfactuals}

Our second target uses the realized outcome as additional information:
\begin{equation}
Y^{(1-a)}\mid X=x,Y^{(a)}=y.
\label{eq:factual-conditioned-law}
\end{equation}
After observing a patient's response to one therapy, this query asks about the response that the same patient would have had under another. Its answer depends on how the two potential outcomes are related, not only on their separate distributions. For example, knowing the response distribution under each therapy does not establish whether a patient with an unusually favorable response to one would also respond favorably to the other. A model for this correspondence specifies the dependence between potential outcomes and thereby determines the factual-conditioned prediction.

\subsection{Identification from Observational Data}

We adopt the standard assumptions for identifying treatment-specific outcome distributions \citep{ma2024diffpo,melnychuk2023normalizing,yoon2018ganite}:
\begin{assumption}\leavevmode
\begin{enumerate}
\item {\it Consistency:} If $A=a$, the observed outcome is $Y=Y^{(a)}$.
\item {\it Unconfoundedness:}
$\{Y^{(0)},Y^{(1)}\}\perp\!\!\!\perp A\mid X$.
\item {\it Overlap:} $0<P(A=1\mid X=x)<1$ for $P_X$-almost every $x$.
\end{enumerate}
\label{assumption 1}
\end{assumption}
Consistency connects the observed response to the potential outcome under the received treatment. Unconfoundedness requires that conditioning on $X$ removes treatment-selection dependence on the potential outcomes. Overlap ensures that both treatments are represented in the population at the covariate values of interest. Together, these assumptions give
\[
p(Y^{(a)}\mid X=x)=p(Y\mid X=x,A=a).
\]
We can therefore learn each potential-outcome distribution using factual observations from its treatment arm.

This equality identifies the two conditional marginals, but leaves their dependence unspecified. The same marginals can arise from different pairings across treatments and hence yield different answers to~\eqref{eq:factual-conditioned-law}. PO-Flow represents a candidate pairing through its invertible maps. Sections~\ref{sec:cf-theory} and~\ref{sec:multivariate-identification} state the additional structural conditions under which this pairing recovers the individual counterfactual.

\section{Proposed Method}
\label{section:algorithm}

PO-Flow learns a treatment-conditioned transformation between outcomes and a common reference distribution. This gives two ways to use the model: drawing a reference sample generates a potential outcome, while encoding a factual outcome and decoding under another treatment gives a candidate counterfactual. We first describe how the maps are trained, then explain these two prediction procedures.

\subsection{Conditional Flow Matching}

The velocity field is conditioned on covariates and treatment:
\begin{equation}
\frac{dy(t)}{dt}=v_\theta(y(t),t;x,a),\qquad t\in[0,1].
\label{our ODE equation}
\end{equation}
We place outcomes at $t=0$ and reference samples $z\sim q=\mathcal N(0,I_{d_Y})$ at $t=1$. Under standard ODE regularity conditions, integration defines an invertible encoder $\Phi_\theta(\cdot;x,a)$ and its decoder $\Phi_\theta^{-1}(\cdot;x,a)$. The treatment-specific maps share the neural parameterization and the reference law, while allowing different outcome distributions.

For each factual tuple $(y^{(a)},x,a)$, draw an independent $z\sim q$ and $t\sim\mathcal U[0,1]$. The linear reference path $\phi_t=(1-t)y^{(a)}+tz$ has velocity $z-y^{(a)}$. Conditional flow matching trains the network to predict this velocity:
\begin{equation}
\mathcal L_{\mathrm{CFM}}(\theta)
=\mathbb E\left[
\left\|v_\theta(\phi_t,t;x,a)-(z-y^{(a)})\right\|_2^2
\right],
\label{our FM loss}
\end{equation}
where the expectation is over factual observations, reference samples, and flow time. Each training update is therefore a regression step along a specified path; it does not require simulating the learned ODE. 

\subsection{Inverse-Probability Weighting}
\label{sec:ipw-method}

The observational loss averages over the treatment assignments present in the data. If treatment is uncommon for some covariates, those treated observations contribute little to the total loss, even though predictions for that treatment may be needed at test time. IPW increases their contribution relative to observations from the more prevalent arm.

Let $e(x)=P(A=1\mid X=x)$. We estimate propensities by $K$-fold: each observation receives a propensity estimate from a model fitted without its fold. The raw weight is
\begin{equation}
r_i=\frac{a_i}{\widehat e_i}
    +\frac{1-a_i}{1-\widehat e_i}.
\label{eq:raw-ipw}
\end{equation}
For instance, when $e(x)=0.1$, a treated observation receives ideal weight $10$, whereas a control observation receives weight $1/0.9$. This compensates for their different frequencies at the same covariate value.

The population interpretation makes this correction explicit. Write $R_a(x)=\mathbb E[\ell_\theta\mid X=x,A=a]$, where $\ell_\theta$ is the flow-matching squared error in~\eqref{our FM loss}. With the true propensity, overlap, and finite expected loss, the normalized ideal weight
\[
W^*=\frac12\left\{\frac{A}{e(X)}+\frac{1-A}{1-e(X)}\right\}
\]
satisfies
\begin{equation}
\mathbb E[W^*\ell_\theta]
=\frac12\mathbb E_X[R_0(X)+R_1(X)].
\label{eq:ipw-population-target}
\end{equation}
Ideal IPW thus gives both arms equal weight at each covariate value while retaining the population covariate distribution. For an unrestricted conditional model, positive weights depending only on $(X,A)$ do not change the pointwise population optimum. Their practical role is to change how a finite-capacity model allocates fitting error across treatment--covariate regions.

Estimated inverse weights can be unstable under weak overlap. We clip the estimated propensities, truncate large weights, normalize them to mean one, and shrink them toward uniform weights. The resulting weights $w_i$ define
\begin{equation}
\widehat{\mathcal L}_{\mathrm{IPW}}(\theta)
=\frac1n\sum_{i=1}^n w_i\,
\mathbb E_{t,z}\left[
\left\|v_\theta(\phi_{i,t},t;x_i,a_i)-(z-y_i)\right\|_2^2
\right].
\label{eq:ipw-cfm}
\end{equation}
Appendix~\ref{appendix:selection-protocol} gives the weight construction and final settings. Regularization trades the exact ideal correction for more stable training. Neither weighting nor cross-fitting supplies outcomes where overlap is absent, or determines the pairing needed for individual counterfactual recovery.

In ACIC16-Vector, we also use a factual mean-regression term with a shared-network readout; Appendix~\ref{appendix:auxiliary-mean} specifies this addition and its use in evaluation.

\subsection{Encoding and Decoding via CNFs}

For a factual outcome $y^{(a)}$, integrating~\eqref{our ODE equation} forward gives
\begin{equation}
z=\Phi_\theta(y^{(a)};x,a)
=y^{(a)}+\int_0^1v_\theta(y(t),t;x,a)\,dt,
\label{forward equation}
\end{equation}
where $y(0)=y^{(a)}$. Conversely, a trajectory with terminal value $\widetilde y(1)=z$ gives
\begin{equation}
\widehat y^{(a)}=\Phi_\theta^{-1}(z;x,a)
=z-\int_0^1v_\theta(\widetilde y(t),t;x,a)\,dt.
\label{reverse process}
\end{equation}
These equations describe the exact flow maps; numerical integration approximates them in implementation.

\emph{Potential-outcome generation} starts with a fresh draw $z\sim q$ and decodes it under treatment $a$. Repeated draws approximate the treatment-specific outcome distribution and, when its first moment exists, its conditional mean. Differences between the two generated means provide a CATE estimate. Algorithm~\ref{alg:interventional} summarizes generation.

\emph{Factual-conditioned counterfactual prediction} instead starts from the observed outcome. We encode $y^{(a)}$ under its factual treatment, hold the resulting reference value fixed, and decode under $1-a$:
\[
\widehat y_{\mathrm{cf}}^{(1-a)}
=\Phi_\theta^{-1}\!\left(\Phi_\theta(y^{(a)};x,a);x,1-a\right).
\]
The retained reference value carries the factual information through the treatment change. It represents a proposed cross-treatment correspondence; a common reference distribution alone does not establish that it preserves the individual's true exogenous noise. Section~\ref{section:theory} gives sufficient conditions for that interpretation. For a fixed factual tuple, this construction returns one outcome rather than sampling a generally nondegenerate factual-conditioned law. Algorithm~\ref{alg:counterfactual} gives the procedure.

\begin{algorithm}[t]
    \caption{Potential-outcome generation with PO-Flow}
    \label{alg:interventional}
    \begin{algorithmic}[1]
        \REQUIRE Trained conditional velocity field $v_{\theta}$
        \INPUT Covariates $x$ and intervention $a\in\{0,1\}$
        \STATE Draw $z\sim q$ at flow time $t=1$.
        \STATE Integrate $\dot y(t)=v_{\theta}(y(t),t;x,a)$ backward from
        $y(1)=z$ to $t=0$.
        \STATE \textbf{Return:} $\hat y^{(a)}:=y(0)$.
    \end{algorithmic}
\end{algorithm}

\begin{algorithm}[t]
    \caption{Factual-conditioned counterfactual prediction with PO-Flow}
    \label{alg:counterfactual}
    \begin{algorithmic}[1]
        \REQUIRE Trained conditional velocity field $v_{\theta}$
        \INPUT Factual observation $(y^{(a)},x,a)$
        \STATE \textbf{Abduction:} integrate
        $\dot y(t)=v_{\theta}(y(t),t;x,a)$ forward from
        $y(0)=y^{(a)}$ to $t=1$, and set $z:=y(1)$.
        \STATE Set the intervention to $\gamma:=1-a$.
        \STATE \textbf{Action--prediction:} integrate
        $\dot{\tilde y}(t)=v_{\theta}(\tilde y(t),t;x,\gamma)$ backward from
        $\tilde y(1)=z$ to $t=0$.
        \STATE \textbf{Return:}
        $\hat y_{\mathrm{cf}}^{(\gamma)}:=\tilde y(0)$.
    \end{algorithmic}
\end{algorithm}

\section{Theoretical Properties}
\label{section:theory}

Encoding and decoding give a prediction procedure, but when does changing treatment in the decoder preserve the individual? For a mechanism $Y^{(a)}=f_X(a,U)$, the requirement is that both potential outcomes of the same individual encode to the same reference value:
\[
\Phi_{x,0}(f_{x,0}(u))=\Phi_{x,1}(f_{x,1}(u)),
\qquad \Phi_{x,a}(y):=\Phi_\theta(y;x,a).
\]
If this equality holds, decoding under either treatment retains the same $u$. Agreement of the encoded distributions alone is weaker: it permits maps that assign different reference values to the same individual. We establish conditions that remove this ambiguity, first through scalar ordering and then through multivariate transport structure. These are sufficient recovery conditions, rather than properties automatically enforced by training. We also derive the CNF formula for potential-outcome density evaluation.

\subsection{Scalar Identification Under a Common Orientation}
\label{sec:cf-theory}

For scalar noise, a common ordering provides a way to align individuals across treatments. If both noise-to-latent maps are increasing, they preserve the conditional rank of $U$. Matching their output distributions then fixes the same latent value for every rank. We formalize this argument using the structural causal model $Y^{(A)}=f_X(A,U)$ with scalar exogenous noise $U$. Write
$P_{U\mid x}$ for the conditional distribution of $U$ given $X=x$, and encode
$Z:=\Phi_\theta(Y^{(A)};X,A)$ via~\eqref{forward equation}. We then assume the following conditions.

\begin{assumption}[Scalar common-orientation condition]\leavevmode
\begin{enumerate}
\item[(A1)] $U \perp\!\!\!\perp A\mid X$. For $P_X$-almost every $x$,
$G_x(u):=P(U\leq u\mid X=x)$ is continuous and strictly increasing on the conditional
support $\mathcal U_x$ of $U\mid X=x$.
\item[(A2)] For each $a\in\{0,1\}$ and $P_X$-almost every $x$,
$m_{x,a}(u):=\Phi_\theta(f_X(a,u);x,a)$ is a continuously differentiable, strictly
increasing bijection from $\mathcal U_x$ onto its image.
\item[(A3)] $Z=\Phi_\theta(Y^{(A)};X,A)\perp\!\!\!\perp A \mid X$.
\end{enumerate}
\label{assumptions}
\end{assumption}

\begin{remark}
Assumption (A1) gives both treatment groups the same conditional distribution of \(U\), while exact training gives (A3) because both encoders map to the same base distribution \(q\). But matching distributions alone does not fix how values of \(U\) correspond across treatments. Assumption (A2) removes this ambiguity by requiring both treatment maps to preserve the same ordering of \(U\). Otherwise, the latent distributions can still match while cross-treatment decoding fails. This holds for additive SCMs \(Y^{(A)}=f_X^*(A)+U\) and more general nonlinear identifiable models \citep{zhang2012identifiability,strobl2023identifying}. If both maps are decreasing, reversing \(U\) reduces to the increasing case.
\end{remark}

\begin{proposition}[Encoded $Z$ as a function of exogenous noise $U$]
\label{thm:encode_as_u}
Let Assumptions~\ref{assumption 1} and~\ref{assumptions} hold. Then there is a
treatment-independent family of continuously differentiable bijections
$\psi_x:\mathcal U_x\to\psi_x(\mathcal U_x)$, defined for $P_X$-almost every $x$, such that
\[
Z=\Phi_\theta(Y^{(A)};X,A)
=\Phi_\theta(f_X(A,U);X,A)
=\psi_X(U).
\]
\end{proposition}
\begin{proof}
Fix an $x$ for which (A1)--(A3) hold. By (A1), $U\mid X=x,A=a$ has distribution
$P_{U\mid x}$ for both treatments. Since $Z=m_{x,a}(U)$, (A3) states that
$m_{x,0}(U)$ and $m_{x,1}(U)$ have the same conditional CDF; denote it by
$F_{Z\mid x}$. For $u\in\mathcal U_x$, strict monotonicity (A2) gives
\begin{equation}
F_{Z\mid x}(m_{x,a}(u))
=\Pr_{U\sim P_{U\mid x}}\{m_{x,a}(U)\leq m_{x,a}(u)\}
=\Pr_{U\sim P_{U\mid x}}\{U\leq u\}=G_x(u).
\end{equation}
By (A1)--(A2), $F_{Z\mid x}$ is continuous and strictly increasing on its support, so
$m_{x,a}(u)=F_{Z\mid x}^{-1}(G_x(u))$ for $a=0,1$. The right-hand side is independent
of $a$; define $\psi_x=m_{x,0}=m_{x,1}$. Then $Z=\psi_X(U)$, and the differentiability
and bijectivity of $\psi_x$ follow from (A2).
\end{proof}

\begin{corollary}[Counterfactual recovery under monotone SCMs]
\label{cor:cf}
Assume Assumptions~\ref{assumption 1} and~\ref{assumptions} hold. Let
$Y^{(A)}=f_X(A,U)$ and $Z=\Phi_\theta(Y^{(A)};X,A)$. Under
$\mathrm{do}(A=\gamma)$, hold $X$ and $U$ fixed. The true counterfactual is
$Y_{\mathrm{cf}}^{(\gamma)}=f_X(\gamma,U)$, and the PO-Flow decoder recovers it almost
surely:
\[
\hat{Y}_{\mathrm{cf}}^{(\gamma)}
:= \Phi_\theta^{-1}(Z;X,\gamma)
= Y_{\mathrm{cf}}^{(\gamma)}.
\]
\end{corollary}
\begin{proof}
Proposition~\ref{thm:encode_as_u} shows that, for the same $X$ and $U$, both treatment
arms encode their outcomes to the common latent value $\psi_X(U)$. Thus the factual latent
value satisfies $Z=\Phi_\theta(f_X(A,U);X,A)=\psi_X(U)=\Phi_\theta(f_X(\gamma,U);X,\gamma)$. Because the
treatment-$\gamma$ encoder is invertible, decoding $Z$ under treatment $\gamma$ gives
$\Phi_\theta^{-1}(Z;X,\gamma)=f_X(\gamma,U)$ almost surely.
\end{proof}

Corollary~\ref{cor:cf} establishes scalar recovery because monotonicity determines a unique map from the scalar noise \(U\) to the latent space. In more than one dimension, vectors have no natural ordering, so scalar monotonicity no longer determines a unique map. The next subsection presents two sufficient ways to resolve this ambiguity: a lower-triangular map that applies the scalar quantile argument coordinate by coordinate in a fixed order, and an order-free convex-gradient map defined by quadratic-cost optimal transport.

\subsection{Identification for Multivariate Outcomes}
\label{sec:multivariate-identification}

Now let $U,Y,Z\in\mathbb R^{d_Y}$, write $f_{x,a}(u)=f_X(a,u)$, and define the
noise-to-latent map
\begin{equation}
m_{x,a}=\Phi_{x,a}\circ f_{x,a},\qquad
\Phi_{x,a}(y):=\Phi_\theta(y;x,a).
\end{equation}
For any vector $v$, let $v_{1:k}=(v_1,\ldots,v_k)$, and let $m_{x,a,k}$ denote the
$k$th coordinate of $m_{x,a}$.
A triangular map extends scalar rank alignment one coordinate at a time: it matches the first coordinate, then the second conditional on the first, and continues in a common order. This accommodates dependence between noise coordinates. We impose the following conditions.

\begin{assumption}[Triangular transport condition]
\label{ass:triangular-transport}
For $P_X$-almost every $x$:
\begin{enumerate}
\item[(B1)] $U\perp\!\!\!\perp A\mid X$, and $P_{U\mid x}$ is absolutely continuous,
with a strictly positive density on its support. For each $k$, the CDF of $U_k$ conditional
on $U_{1:k-1}$ and $X=x$ is continuous and strictly increasing;
\item[(B2)] $f_{x,a}$ maps the noise support bijectively onto the outcome support almost
everywhere, and the flow encoder $\Phi_{x,a}$ is invertible;
\item[(B3)] both $m_{x,a}$ are lower triangular in a common coordinate order,
$m_{x,a,k}(u)=m_{x,a,k}(u_{1:k})$, and, for almost every prefix $u_{1:k-1}$,
$u_k\mapsto m_{x,a,k}(u_{1:k})$ is continuous and strictly increasing;
\item[(B4)] under both treatments, the encoded latent variable has the same reference
distribution $Q_x$: $(m_{x,a})_\#P_{U\mid x}=Q_x$ for $a\in\{0,1\}$. For $Z\sim Q_x$,
the CDF of $Z_k$ conditional on $Z_{1:k-1}$ is continuous and strictly increasing.
\end{enumerate}
\end{assumption}

Conditions (B1)--(B2) ensure that the conditional quantiles and inverse maps are
well defined. Condition (B4) makes both treatment maps produce the same latent
distribution, but does not determine which latent value corresponds to each $u$.
Condition (B3) removes this ambiguity by applying the same sequence of conditional
quantile maps in a fixed coordinate order, following the classical triangular
construction \citep{rosenblatt1952remarks,knothe1957contributions}.
Theorem~\ref{thm:multivariate-transport} uses this uniqueness to prove that the two
treatment maps agree and that PO-Flow recovers the individual counterfactual.

\begin{theorem}[Multivariate shared-noise identification]
\label{thm:multivariate-transport}
Under Assumption~\ref{ass:triangular-transport}, there is a unique lower-triangular,
diagonally increasing map $\psi_x$ transporting $P_{U\mid x}$ to $Q_x$, and
\begin{equation}
m_{x,0}(u)=m_{x,1}(u)=\psi_x(u)
\quad P_{U\mid x}\text{-almost surely}.
\label{eq:shared-multivariate-map}
\end{equation}
Consequently, for every factual treatment $a$ and intervention $\gamma$,
\begin{equation}
\Phi_{x,\gamma}^{-1}\!\left(\Phi_{x,a}(f_{x,a}(u))\right)
=f_{x,\gamma}(u)
\quad P_{U\mid x}\text{-almost surely}.
\label{eq:multivariate-cf-recovery}
\end{equation}
\end{theorem}

\begin{proof}
We prove uniqueness coordinate by coordinate. For the first coordinate, triangularity and
strict increase imply the quantile identity
\begin{equation}
m_{x,a,1}(u_1)=F^{-1}_{Q_{x,1}}\!\left(F_{U_1\mid x}(u_1)\right),
\end{equation}
so the first coordinate is independent of $a$. Assume inductively that
$m_{x,0,j}=m_{x,1,j}$ for every $j<k$, and call this shared function $\psi_{x,j}$. Let
$\psi_{x,1:k-1}=(\psi_{x,1},\ldots,\psi_{x,k-1})$. This map is invertible on its image,
so $Z_{1:k-1}=\psi_{x,1:k-1}(u_{1:k-1})$ uniquely determines
$U_{1:k-1}=u_{1:k-1}$. By (B4), the increasing $k$th coordinate is therefore the
conditional quantile map:
\begin{equation}
m_{x,a,k}(u_{1:k})
=F^{-1}_{Q_{x,k}\mid Z_{1:k-1}=\psi_{x,1:k-1}(u_{1:k-1})} \left(F_{U_k\mid U_{1:k-1}=u_{1:k-1},X=x}(u_k)\right).
\label{eq:conditional-transport}
\end{equation}
The right-hand side is independent of treatment, so induction establishes
\eqref{eq:shared-multivariate-map}. Finally,
$\Phi_{x,\gamma}(f_{x,\gamma}(u))=\psi_x(u)=
\Phi_{x,a}(f_{x,a}(u))$.  Applying the inverse of $\Phi_{x,\gamma}$ proves
\eqref{eq:multivariate-cf-recovery}.
\end{proof}

The triangular construction depends on a coordinate order. An alternative is to select the noise-to-latent map through quadratic-cost optimal transport. When both maps are gradients of convex functions, transport uniqueness forces them to agree. The next corollary uses this property to obtain the same recovery conclusion without choosing a coordinate order.

\begin{corollary}[Convex-gradient alternative]
\label{cor:brenier}
Keep (B1), (B2), and (B4), but replace (B3) with the following condition:
$P_{U\mid x}$ and $Q_x$ have finite second moments, and each $m_{x,a}$ is
the gradient of a convex function. Then
$m_{x,0}=m_{x,1}$ $P_{U\mid x}$-almost surely, and
\eqref{eq:multivariate-cf-recovery} holds.
\end{corollary}
\begin{proof}
By (B1), $P_{U\mid x}$ is absolutely continuous. Brenier's theorem \citep{brenier1991polar,villani2009optimal} states
that, under the finite-second-moment condition, there is a unique
$P_{U\mid x}$-almost-everywhere map that is the gradient of a convex
function and transforms $P_{U\mid x}$ into $Q_x$. By (B4), both $m_{x,0}$ and
$m_{x,1}$ perform this transformation, and by the replacement condition
both have the required convex-gradient form. Hence
$m_{x,0}=m_{x,1}$ almost everywhere. Since
$m_{x,a}=\Phi_{x,a}\circ f_{x,a}$, this equality gives
$\Phi_{x,\gamma}(f_{x,\gamma}(u))
=\Phi_{x,a}(f_{x,a}(u))$. Applying the inverse of $\Phi_{x,\gamma}$,
which exists by (B2), proves \eqref{eq:multivariate-cf-recovery}.
\end{proof}

Theorem~\ref{thm:multivariate-transport} and
Corollary~\ref{cor:brenier} establish two complementary routes to multivariate
counterfactual recovery. The triangular condition uses a fixed coordinate order,
whereas the convex-gradient condition selects the order-free quadratic-cost optimal
transport. Convex-gradient maps can in principle be parameterized using input-convex
neural networks \citep{amos2017input,makkuva2020optimal}, although this requires a constrained architecture and demanding optimization. Our model
remains unconstrained, so the results give sufficient recovery guarantees when either
structure holds. Proposition~\ref{prop:rotation-counterexample} shows why matching the
latent distributions alone cannot provide the same guarantee.

\begin{proposition}[Latent invariance alone is insufficient]
\label{prop:rotation-counterexample}
For every $d_Y\geq2$, there exist a smooth invertible SCM and smooth invertible encoders
that satisfy $Z\perp\!\!\!\perp A\mid X$ exactly, but cross-treatment decoding does not
recover the individual counterfactual.
\end{proposition}
\begin{proof}
Fix $X=x$. Let $U\sim\mathcal N(0,I_{d_Y})$ and let
$A\sim\operatorname{Bernoulli}(p)$, $0<p<1$, be independent of $U$. Take
$f_{x,0}(u)=f_{x,1}(u)=u$, so a unit with noise $u$ has $Y^{(0)}=Y^{(1)}=u$.
Let $R$ be a nonidentity rotation of two coordinates and set
$\Phi_{x,0}(y)=y$ and $\Phi_{x,1}(y)=Ry$. Then $Z=U$ when $A=0$ and $Z=RU$ when $A=1$.
Since $RU\sim\mathcal N(0,I_{d_Y})$, the two conditional laws of $Z$ are identical, and hence
$Z\perp\!\!\!\perp A\mid X$. However, for a control unit with $Y=u$, cross-treatment decoding gives
$\Phi_{x,1}^{-1}(\Phi_{x,0}(u))=R^{-1}u\neq u=f_{x,1}(u)$ almost surely. A rotation is
generated by a smooth linear ODE, so all maps in this example are smooth and invertible and
the example applies to CNF encoders.
\end{proof}

In this example, treatment has no effect, yet one encoder rotates the individual before the alternative decoder reconstructs the outcome. Every marginal distribution remains correct. The failure is entirely in the correspondence across treatments, which is the ambiguity excluded by the triangular or convex-gradient conditions.

\subsection{Likelihood Evaluation for Potential Outcome Densities}
\label{sec:density-theory}

The same ODE also gives the density of a generated outcome. Its velocity field can expand or contract local volume, and the divergence records that change along a trajectory. Combining this accumulated volume change with the reference density gives the outcome density through the instantaneous change-of-variables formula for CNFs.

\begin{proposition}
\label{log density}
Given a base sample $z\sim q(\cdot)$ at $t=1$, the log-density of the decoded outcome
at $t=0$ is
\begin{equation}
\log p_{\theta,Y\mid X,A}(y\mid x,a)=
\log q(z)
+
\int_{0}^{1}
\nabla_{y}\!\cdot\,v_{\theta}\bigl(y(t),t;\,x,a\bigr)\,dt,
\label{equation log density}
\end{equation}
where $y = \Phi_{\theta}^{-1}(z;x,a)$.
\end{proposition}
\begin{proof}
Let $p_t(\cdot\mid x,a)$ denote the conditional density of $y(t)$. Along the trajectory
$dy(t)/dt=v_\theta(y(t),t;x,a)$, the instantaneous change-of-variables formula gives
$d\log p_t(y(t)\mid x,a)/dt=-\nabla_y\!\cdot v_\theta(y(t),t;x,a)$. Integrating gives
$\log p_1(z\mid x,a)-\log p_0(y\mid x,a)
=-\int_0^1\nabla_y\!\cdot v_\theta(y(t),t;x,a)\,dt$, where $y(0)=y$ and $y(1)=z$.
Since $p_0=p_{\theta,Y\mid X,A}$ and $p_1=q$, moving the integral to the other side gives
\eqref{equation log density}.
\end{proof}

\section{Experiments}

The experiments examine whether reweighting improves the fitted predictors and how the flow construction performs as outcomes become multivariate or image-valued. The scalar ablation isolates the training weights, while the broader comparisons assess the predictions available from each method. Counterfactual errors are interpreted relative to each benchmark's specified pairing of potential outcomes.

\subsection{Datasets and Baselines}

We evaluate four scalar benchmarks---ACIC 2016, ACIC 2018, IHDP, and IBM---and seven
vector or image settings based on IHDP, ACIC16, and MorphoMNIST. For scalar outcomes,
Table~\ref{tab:ipw-stable} compares PO-Flow with ten baselines: TARNet and CFR
\citep{shalit2017estimating}, DragonNet \citep{shi2019adapting}, DiffPO
\citep{ma2024diffpo}, INFs \citep{melnychuk2023normalizing}, S- and T-learners
\citep{kunzel2019metalearners}, the DR-learner \citep{kennedy2023towards}, CEVAE
\citep{louizos2017causal}, and GANITE \citep{yoon2018ganite}. For vector and image outcomes,
we compare multivariate adaptations of TARNet, DragonNet, CFR, CEVAE, and GANITE. Because
paired potential outcomes are not observed in real data, the benchmarks provide
semi-synthetic ground truth. Appendix~\ref{appendix:experimental details}
gives the construction details. Code is available at \url{https://github.com/StatFusion/PO-Flow}.

\subsection{Evaluation Metrics}
\label{sec:evaluation-metrics}

For $n$ test units with $d=d_Y$ outcome coordinates, let $\mu_{i,a}$ denote the conditional mean under treatment $a$, $\widehat\mu_{i,a}$ its estimate, and $\widehat y_{i,\mathrm{cf}}^{(1-a_i)}$ the predicted alternative outcome. We report counterfactual RMSE, conditional treatment-effect error (PEHE), and factual-outcome prediction RMSE:
\begin{equation}
\begin{aligned}
\mathrm{CF\text{-}RMSE}
&=\left[\frac{1}{nd}\sum_{i=1}^n
\|\widehat y_{i,\mathrm{cf}}^{(1-a_i)}-y_i^{(1-a_i)}\|_2^2\right]^{1/2},\\
\mathrm{PEHE}
&=\left[\frac{1}{nd}\sum_{i=1}^n
\|(\widehat\mu_{i,1}-\widehat\mu_{i,0})-(\mu_{i,1}-\mu_{i,0})\|_2^2\right]^{1/2},\\
\mathrm{PO\text{-}RMSE}
&=\left[\frac{1}{nd}\sum_{i=1}^n
\|\widehat\mu_{i,a_i}-y_i^{(a_i)}\|_2^2\right]^{1/2}.
\end{aligned}
\label{eq:evaluation-point-metrics}
\end{equation}
The reported PO-RMSE evaluates the factual arm only. For MorphoMNIST, the supplied references are paired individual images rather than conditional means; its PEHE therefore measures paired image-effect prediction error. CF-RMSE uses factual-conditioned predictions where supported and covariate-only predictions for TARNet, DragonNet, and CFR.

Let $G_a^{(m)}=\{\widehat y_{i,a}^{(m)}\}_{i=1}^n$, $G_a^\mu=\{\widehat\mu_{i,a}\}_{i=1}^n$, and $R_a=\{\mu_{i,a}\}_{i=1}^n$. With $\mathcal W$ denoting scalar Wasserstein distance or its sliced multivariate counterpart, the reported discrepancies are
\begin{equation}
D_{\mathrm{sample}}
=\frac{1}{2M}\sum_{a=0}^1\sum_{m=1}^M\mathcal W(G_a^{(m)},R_a),
\qquad
D_{\mathrm{mean}}
=\frac12\sum_{a=0}^1\mathcal W(G_a^\mu,R_a).
\label{eq:evaluation-distribution-metrics}
\end{equation}
Tables label these scores $W_1$ and $W_{1,\mu}$ for scalar outcomes, and $\operatorname{SW}_1$ and $\operatorname{SW}_{1,\mu}$ otherwise. Both use the supplied mean references, so the sample score does not directly measure fit to the noisy potential-outcome distribution. Scalar errors use original units; vector coordinates use training-factual standardization, and image pixels use $[-1,1]$. Appendix~\ref{appendix:selection-protocol} gives the normalization and sampling settings.

\subsection{Scalar Outcomes: Overlap-Aware Evaluation}
\label{sec:ipw-results}

For the scalar benchmarks, propensities $P(A=1\mid X=x)$ are estimated by five-fold cross-fitted logistic
regression on the training split and clipped to $[0.01,0.99]$.
Appendix~\ref{appendix:selection-protocol} gives the remaining weighting and evaluation details.

\textit{Table~\ref{tab:ipw-stable}} presents the scalar results for PO-Flow and all
baselines on ACIC16, ACIC18, IHDP, and IBM. For each available metric (CF-RMSE, PEHE,
PO-RMSE, $W_1$, and $W_{1,\mu}$), entries are reported as the mean with standard
deviation. All the methods are evaluated over seeds 42--46.

\begin{table}[h!]
\centering
\caption{Scalar-outcome comparison over five seeds (mean with standard
deviation); lower is better.}
\label{tab:ipw-stable}
\scriptsize
\setlength{\tabcolsep}{2.2pt}
\begin{tabular}{llccccc}
\toprule
\textbf{Data} & \textbf{Method} & \textbf{CF-RMSE} & \textbf{PEHE} & \textbf{PO-RMSE} & $\mathbf{W_1}$ & $\mathbf{W_{1,\mu}}$\\
\midrule
ACIC16 & \textbf{PO-Flow} & \mstd{1.52}{.01} & \textbf{\mstd{.41}{.02}} & \textbf{\mstd{1.25}{.04}} & \textbf{\mstd{.17}{.01}} & \textbf{\mstd{.17}{.02}}\\
 & PO-Flow (no IPW) & \mstd{1.53}{.17} & \mstd{.82}{.09} & \mstd{1.37}{.09} & \mstd{.42}{.12} & \mstd{.45}{.10}\\
 & TARNet & \mstd{1.53}{.10} & \mstd{.58}{.18} & \mstd{1.26}{.08} & -- & \mstd{.27}{.10}\\
 & DragonNet & \textbf{\mstd{1.34}{.12}} & \mstd{.60}{.14} & \mstd{1.27}{.08} & -- & \mstd{.22}{.08}\\
 & DiffPO & \mstd{1.78}{.20} & \mstd{1.12}{.26} & \mstd{1.42}{.08} & \mstd{.51}{.18} & --\\
 & INFs & -- & \mstd{1.42}{.30} & \mstd{1.36}{.25} & \mstd{.67}{.14} & --\\
 & S-learner & -- & \mstd{3.87}{.98} & \mstd{1.86}{.22} & -- & --\\
 & T-learner & -- & \mstd{5.15}{.87} & \mstd{1.94}{.28} & -- & --\\
 & DR-learner & -- & \mstd{3.51}{.23} & -- & -- & --\\
 & CEVAE & -- & \mstd{2.31}{.74} & \mstd{2.06}{.40} & \mstd{.70}{.21} & --\\
 & CFR & \mstd{3.38}{.30} & \mstd{2.74}{.60} & \mstd{1.57}{.11} & -- & --\\
 & GANITE & \mstd{3.67}{.55} & \mstd{3.11}{.74} & \mstd{1.46}{.23} & \mstd{.86}{.23} & --\\
\addlinespace[1pt]
ACIC18 & \textbf{PO-Flow} & \textbf{\mstd{.01}{.00}} & \textbf{\mstd{.01}{.00}} & \textbf{\mstd{.43}{.01}} & \textbf{\mstd{.12}{.03}} & \textbf{\mstd{.30}{.00}}\\
 & PO-Flow (no IPW) & \mstd{.05}{.02} & \mstd{.07}{.02} & \mstd{.60}{.12} & \mstd{.12}{.04} & \mstd{.38}{.03}\\
 & TARNet & \mstd{.56}{.01} & \mstd{.52}{.02} & \mstd{.55}{.01} & -- & \mstd{.05}{.01}\\
 & DragonNet & \mstd{.59}{.02} & \mstd{.57}{.01} & \mstd{.57}{.00} & -- & \textbf{\mstd{.04}{.01}}\\
 & DiffPO & \mstd{.65}{.15} & \mstd{.08}{.04} & \mstd{.79}{.16} & \mstd{.13}{.05} & --\\
 & INFs & -- & \mstd{.37}{.12} & \mstd{.78}{.17} & \mstd{.16}{.05} & --\\
 & S-learner & -- & \mstd{.07}{.03} & \mstd{.57}{.11} & -- & --\\
 & T-learner & -- & \mstd{1.32}{.53} & \mstd{1.71}{.40} & -- & --\\
 & DR-learner & -- & \mstd{.57}{.10} & -- & -- & --\\
 & CEVAE & -- & \mstd{1.21}{.16} & \mstd{.85}{.17} & \mstd{1.02}{.23} & --\\
 & CFR & \mstd{1.01}{.07} & \mstd{.80}{.18} & \mstd{.98}{.12} & -- & --\\
 & GANITE & \mstd{.70}{.14} & \mstd{.60}{.13} & \mstd{.97}{.15} & \mstd{1.18}{.25} & --\\
\addlinespace[1pt]
IHDP & \textbf{PO-Flow} & \textbf{\mstd{1.57}{.13}} & \textbf{\mstd{.39}{.07}} & \textbf{\mstd{1.05}{.05}} & \mstd{.45}{.03} & \textbf{\mstd{.13}{.03}}\\
 & PO-Flow (no IPW) & \mstd{1.69}{.10} & \mstd{.45}{.10} & \mstd{1.19}{.13} & \textbf{\mstd{.41}{.06}} & \mstd{.14}{.04}\\
 & TARNet & \textbf{\mstd{1.58}{.07}} & \mstd{1.44}{.06} & \mstd{1.32}{.11} & -- & \mstd{.42}{.05}\\
 & DragonNet & \mstd{1.50}{.13} & \mstd{1.51}{.19} & \mstd{1.36}{.09} & -- & \mstd{.50}{.08}\\
 & DiffPO & \mstd{1.85}{.23} & \mstd{.84}{.18} & \mstd{1.40}{.22} & \mstd{.88}{.17} & --\\
 & INFs & -- & \mstd{.95}{.27} & \mstd{1.20}{.12} & \textbf{\mstd{.40}{.09}} & --\\
 & S-learner & -- & \mstd{.93}{.08} & \mstd{1.44}{.20} & -- & --\\
 & T-learner & -- & \mstd{1.31}{.09} & \mstd{1.49}{.27} & -- & --\\
 & DR-learner & -- & \mstd{1.28}{.15} & -- & -- & --\\
 & CEVAE & -- & \mstd{1.43}{.10} & \mstd{1.36}{.18} & \mstd{.76}{.20} & --\\
 & CFR & \mstd{1.65}{.25} & \mstd{.78}{.04} & \mstd{1.18}{.20} & -- & --\\
 & GANITE & \mstd{2.72}{.59} & \mstd{2.40}{.40} & \mstd{1.67}{.34} & \mstd{.82}{.29} & --\\
\addlinespace[1pt]
IBM & \textbf{PO-Flow} & \textbf{\mstd{.03}{.01}} & \textbf{\mstd{.03}{.01}} & \textbf{\mstd{.40}{.02}} & \mstd{.47}{.03} & \textbf{\mstd{.23}{.02}}\\
 & PO-Flow (no IPW) & \mstd{.04}{.02} & \mstd{.04}{.01} & \mstd{.97}{.11} & \textbf{\mstd{.44}{.08}} & \mstd{.26}{.04}\\
 & TARNet & \mstd{.53}{.05} & \mstd{.48}{.07} & \mstd{.50}{.03} & -- & \textbf{\mstd{.09}{.02}}\\
 & DragonNet & \mstd{.66}{.07} & \mstd{.65}{.08} & \mstd{.51}{.02} & -- & \mstd{.10}{.05}\\
 & DiffPO & \mstd{.64}{.16} & \mstd{.09}{.03} & \mstd{1.85}{.35} & \mstd{.75}{.14} & --\\
 & INFs & -- & \mstd{.18}{.08} & \mstd{1.57}{.28} & \textbf{\mstd{.45}{.09}} & --\\
 & S-learner & -- & \mstd{.07}{.04} & \mstd{1.16}{.19} & -- & --\\
 & T-learner & -- & \mstd{.69}{.15} & \mstd{1.96}{.50} & -- & --\\
 & DR-learner & -- & \mstd{.19}{.05} & -- & -- & --\\
 & CEVAE & -- & \mstd{.75}{.16} & \mstd{.98}{.12} & \textbf{\mstd{.45}{.07}} & --\\
 & CFR & \mstd{.87}{.16} & \mstd{1.08}{.28} & \mstd{2.17}{.48} & -- & --\\
 & GANITE & \mstd{2.02}{.37} & \mstd{.80}{.24} & \mstd{2.59}{.47} & \mstd{1.34}{.30} & --\\
\bottomrule
\end{tabular}
\end{table}

The controlled ablation in Table~\ref{tab:ipw-stable} isolates the effect of weighting: the architecture and training protocol are unchanged. IPW lowers the reported mean CF-RMSE, PEHE, and PO-RMSE on all four datasets. Its effect differs across targets. On ACIC16, PEHE falls from $0.82$ to $0.41$, while CF-RMSE changes only from $1.53$ to $1.52$. Thus, better estimation of conditional mean effects need not imply a comparable improvement in individual counterfactual prediction. The five-run standard deviations also matter when interpreting these differences.

Across methods, PO-Flow has the lowest reported PEHE and PO-RMSE on all four scalar benchmarks. Counterfactual performance is more mixed: it has the lowest reported CF-RMSE on ACIC18 and IBM, while DragonNet has lower values on ACIC16 and IHDP. These comparisons use the available prediction mechanism for each method, as described in Section~\ref{sec:evaluation-metrics}. The mean-reference Wasserstein scores likewise show no uniform improvement from IPW. Appendix~\ref{appendix:overlap-diagnostics} provides covariate-balance and weight-concentration diagnostics, and Appendix~\ref{appendix:scalar-density} reports the separate Gaussian-reference density experiment.

\subsection{Vector and Image Outcomes}
\label{sec:vector-benchmarks}

Multivariate outcomes test whether the same construction can handle several responses jointly. For the tabular benchmarks, we construct vector-valued outcomes by combining
$d_Y\in\{2,5,10\}$ scalar response surfaces for each unit, while keeping the
covariates and treatment assignment fixed across outcome coordinates. For
MorphoMNIST-Intensity, the two potential outcomes are paired $28\times28$ images
corresponding to low- and high-intensity interventions, giving $d_Y=784$
\citep{castro2019morphomnist,melistas2024benchmarking}. We use $d_Y=5$ as the
primary tabular setting and $d_Y=2$ and $10$ to examine sensitivity to outcome
dimension. Further construction details are given in
Appendix~\ref{appendix:experimental details}.

\begin{table}[h!]
\centering
\caption{Vector- and image-outcome comparison over five seeds (mean with
standard deviation); lower is better.}
\label{tab:vector-results}
\scriptsize
\setlength{\tabcolsep}{2.2pt}
\begin{tabular}{lcrccccc}
\toprule
\textbf{Data} & $\mathbf{d_Y}$ & \textbf{Method} & \textbf{CF-RMSE} & \textbf{PEHE} & \textbf{PO-RMSE} & $\mathbf{SW_1}$ & $\mathbf{SW_{1,\mu}}$\\
\midrule
IHDP & 2 & \textbf{PO-Flow} & \mstd{.69}{.00} & \textbf{\mstd{.20}{.01}} & \textbf{\mstd{.48}{.00}} & \textbf{\mstd{.21}{.00}} & \textbf{\mstd{.05}{.00}}\\
 & & TARNet & \mstd{.64}{.02} & \mstd{.55}{.03} & \mstd{.56}{.01} & -- & \mstd{.16}{.01}\\
 & & DragonNet & \mstd{.64}{.01} & \mstd{.59}{.03} & \mstd{.57}{.01} & -- & \mstd{.19}{.01}\\
 & & CFR & \textbf{\mstd{.63}{.02}} & \mstd{.55}{.03} & \mstd{.56}{.01} & -- & \mstd{.16}{.01}\\
 & & GANITE & \mstd{1.15}{.37} & \mstd{1.03}{.47} & \mstd{.76}{.11} & -- & \mstd{.44}{.21}\\
 & & CEVAE & \mstd{1.03}{.16} & \mstd{.39}{.02} & \mstd{.54}{.01} & \mstd{.46}{.01} & \mstd{.11}{.02}\\
\addlinespace[1pt]
IHDP & 5 & \textbf{PO-Flow} & \textbf{\mstd{.62}{.00}} & \textbf{\mstd{.20}{.00}} & \textbf{\mstd{.44}{.00}} & \textbf{\mstd{.16}{.00}} & \textbf{\mstd{.06}{.00}}\\
 & & TARNet & \mstd{.65}{.02} & \mstd{.44}{.03} & \mstd{.49}{.00} & -- & \mstd{.13}{.01}\\
 & & DragonNet & \mstd{.65}{.02} & \mstd{.45}{.02} & \mstd{.50}{.00} & -- & \mstd{.14}{.01}\\
 & & CFR & \mstd{.64}{.02} & \mstd{.43}{.03} & \mstd{.49}{.01} & -- & \mstd{.13}{.01}\\
 & & GANITE & \mstd{1.51}{.54} & \mstd{1.39}{.56} & \mstd{.82}{.16} & -- & \mstd{.59}{.25}\\
 & & CEVAE & \mstd{.66}{.07} & \mstd{.31}{.02} & \mstd{.55}{.01} & \mstd{.38}{.00} & \mstd{.16}{.01}\\
\addlinespace[1pt]
IHDP & 10 & \textbf{PO-Flow} & \textbf{\mstd{.60}{.00}} & \textbf{\mstd{.22}{.00}} & \textbf{\mstd{.43}{.00}} & \textbf{\mstd{.15}{.00}} & \textbf{\mstd{.06}{.00}}\\
 & & TARNet & \textbf{\mstd{.60}{.01}} & \mstd{.38}{.01} & \mstd{.46}{.00} & -- & \mstd{.12}{.00}\\
 & & DragonNet & \mstd{.62}{.01} & \mstd{.42}{.01} & \mstd{.47}{.01} & -- & \mstd{.14}{.01}\\
 & & CFR & \textbf{\mstd{.60}{.01}} & \mstd{.38}{.01} & \mstd{.46}{.00} & -- & \mstd{.12}{.00}\\
 & & GANITE & \mstd{1.96}{.41} & \mstd{1.87}{.41} & \mstd{.96}{.12} & -- & \mstd{.79}{.19}\\
 & & CEVAE & \mstd{.63}{.03} & \mstd{.47}{.01} & \mstd{.63}{.01} & \mstd{.37}{.00} & \mstd{.23}{.01}\\
\midrule
ACIC16 & 2 & \textbf{PO-Flow} & \mstd{.32}{.00} & \textbf{\mstd{.18}{.01}} & \textbf{\mstd{.27}{.02}} & \textbf{\mstd{.04}{.00}} & \textbf{\mstd{.03}{.00}}\\
 & & TARNet & \mstd{.33}{.00} & \mstd{.24}{.01} & \mstd{.28}{.01} & -- & \mstd{.04}{.01}\\
 & & DragonNet & \mstd{.39}{.03} & \mstd{.33}{.03} & \mstd{.30}{.01} & -- & \mstd{.07}{.01}\\
 & & CFR & \textbf{\mstd{.31}{.00}} & \mstd{.25}{.02} & \mstd{.29}{.01} & -- & \mstd{.05}{.01}\\
 & & GANITE & \mstd{.93}{.08} & \mstd{.88}{.09} & \mstd{.46}{.04} & -- & \mstd{.27}{.07}\\
 & & CEVAE & \mstd{1.06}{.07} & \mstd{.45}{.05} & \mstd{.41}{.05} & \mstd{.08}{.01} & \mstd{.05}{.01}\\
\addlinespace[1pt]
ACIC16 & 5 & \textbf{PO-Flow} & \textbf{\mstd{.39}{.00}} & \textbf{\mstd{.33}{.01}} & \mstd{.31}{.00} & \textbf{\mstd{.03}{.00}} & \textbf{\mstd{.04}{.00}}\\
 & & TARNet & \textbf{\mstd{.39}{.00}} & \mstd{.35}{.01} & \textbf{\mstd{.31}{.01}} & -- & \mstd{.05}{.01}\\
 & & DragonNet & \mstd{.44}{.01} & \mstd{.42}{.02} & \mstd{.33}{.00} & -- & \mstd{.06}{.01}\\
 & & CFR & \mstd{.40}{.00} & \mstd{.36}{.02} & \mstd{.32}{.01} & -- & \mstd{.04}{.01}\\
 & & GANITE & \mstd{1.13}{.35} & \mstd{1.08}{.42} & \mstd{.53}{.10} & -- & \mstd{.31}{.17}\\
 & & CEVAE & \mstd{.87}{.04} & \mstd{.67}{.04} & \mstd{.65}{.05} & \mstd{.08}{.01} & \mstd{.08}{.01}\\
\addlinespace[1pt]
ACIC16 & 10 & \textbf{PO-Flow} & \mstd{.50}{.01} & \mstd{.55}{.01} & \mstd{.47}{.01} & \textbf{\mstd{.05}{.00}} & \mstd{.06}{.00}\\
 & & TARNet & \textbf{\mstd{.45}{.01}} & \textbf{\mstd{.42}{.01}} & \textbf{\mstd{.35}{.00}} & -- & \textbf{\mstd{.05}{.01}}\\
 & & DragonNet & \mstd{.52}{.01} & \mstd{.51}{.01} & \mstd{.39}{.00} & -- & \mstd{.07}{.01}\\
 & & CFR & \mstd{.45}{.01} & \mstd{.42}{.01} & \mstd{.35}{.00} & -- & \mstd{.05}{.01}\\
 & & GANITE & \mstd{1.33}{.12} & \mstd{1.28}{.13} & \mstd{.61}{.04} & -- & \mstd{.40}{.05}\\
 & & CEVAE & \mstd{1.13}{.05} & \mstd{.73}{.06} & \mstd{.76}{.06} & \mstd{.10}{.01} & \mstd{.37}{.01}\\
\midrule
MorphoMNIST & 784 & \textbf{PO-Flow} & \textbf{\mstd{.13}{.00}} & \textbf{\mstd{.18}{.00}} & \mstd{.35}{.00} & \mstd{.29}{.00} & \textbf{\mstd{.09}{.00}}\\
 & & TARNet & \mstd{.31}{.00} & \mstd{.20}{.00} & \mstd{.31}{.00} & -- & \mstd{.14}{.00}\\
 & & DragonNet & \mstd{.30}{.00} & \mstd{.20}{.00} & \textbf{\mstd{.31}{.00}} & -- & \mstd{.14}{.00}\\
 & & CFR & \mstd{.31}{.00} & \mstd{.20}{.00} & \mstd{.31}{.00} & -- & \mstd{.15}{.00}\\
 & & GANITE & \mstd{.22}{.02} & \mstd{.24}{.01} & \mstd{.32}{.00} & -- & \mstd{.16}{.00}\\
 & & CEVAE & \mstd{.37}{.00} & \mstd{.22}{.00} & \mstd{.35}{.00} & \textbf{\mstd{.09}{.00}} & \mstd{.26}{.00}\\
\bottomrule
\end{tabular}
\end{table}

For the vector and image experiments, we implement PyTorch adaptations of TARNet,
DragonNet, CFR, GANITE, and CEVAE with network sizes matched to PO-Flow, using the same
data splits, seeds, and training-only preprocessing. TARNet, DragonNet, CFR, and GANITE are point-prediction methods in these experiments and
therefore do not provide distributional $\operatorname{SW}_1$. However, PEHE, PO-RMSE, and
$\operatorname{SW}_{1,\mu}$ are evaluated consistently across all methods. Network specifications and adaptation details are provided in
Appendices~\ref{appendix: nn structure and implementation},
\ref{appendix:overlap-diagnostics}, and~\ref{appendix:baseline-adaptations}.

Table~\ref{tab:vector-results} shows that performance depends on both the dataset and the prediction target. PO-Flow has lower reported mean-effect and factual-outcome errors than the comparators in the three IHDP settings, but its counterfactual error is higher at $d_Y=2$. In ACIC16, increasing the dimension to ten reverses the advantage on several point-prediction metrics: TARNet and CFR outperform PO-Flow. Higher output dimension therefore does not by itself favor the generative model.

The image setting illustrates a different distinction. The reported encode--decode counterfactual error is lower than that of the comparison methods, while factual-outcome prediction error is higher than that of several point predictors. Conditioning on an observed image and predicting an image from covariates alone are different tasks, and success on one does not establish success on the other. The image-effect and mean-reference distance targets retain the qualifications stated in Section~\ref{sec:evaluation-metrics}.

\section{Discussion and Conclusions}

We extend PO-Flow by incorporating cross-fitted inverse-probability
weighting into flow-matching training and establishing sufficient
conditions for multivariate counterfactual recovery. The framework
represents treatment-specific outcome distributions through invertible
maps, supporting potential-outcome generation, CATE estimation,
factual-conditioned counterfactual prediction, and density evaluation.
Experiments on scalar, multivariate, and image outcomes examine these
tasks and the effect of weighted training. The central distinction is
between learning the outcome distributions and recovering the individual
correspondence across treatments: IPW addresses the allocation of training
error under observational assignment, while counterfactual recovery
requires additional structural assumptions.

These results have several limitations. Individual counterfactual
outcomes are generally unavailable in observational data, so direct
evaluation requires a specified simulator or another source of valid
paired counterfactual ground truth. Such evaluations depend on how the
benchmark couples the potential outcomes and do not by themselves
establish recovery in real observational settings. IPW also depends on
the quality of propensity estimates and adequate overlap. Clipping and
shrinkage can stabilize training, but modify the ideal weighted objective
and cannot supply information about treatments absent from a covariate
region.

Our recovery theory describes an idealized setting with exact agreement
of the encoded distributions. For multivariate outcomes, it additionally
requires the noise-to-latent maps to have a common triangular monotone
structure or a convex-gradient form. These restrictions are not enforced
by the current training objective. The results therefore explain when
encode--decode prediction recovers an individual counterfactual, rather
than guaranteeing recovery for every fitted model. Future work can study
architectures that impose these restrictions and quantify how
distribution-fitting and numerical integration errors affect
counterfactual predictions.

\section*{Acknowledgment}
The work of D.W. and Y.X. is supported by NSF (CMMI-2112533) and the Coca-Cola Foundation. The work of D.I. is supported by NSF (IIS-2212097) and ARL (W911NF-2020-221).

\appendices

\setlength{\parindent}{0pt}
\setlength{\parskip}{0pt}

\section{Experimental Details}
\label{appendix:experimental details}

This section specifies the benchmark construction, configurations,
preprocessing, and implementation choices used for the reported results.

\subsection{Benchmarks and Ground Truth}
\label{data synthetic}
Because individual counterfactuals are unobserved, we use semi-synthetic benchmarks with known response surfaces: ACIC 2016 (4,802 units, 82 covariates), ACIC 2018 (1,000, 177), IHDP (747, 25) \citep{hill2011bayesian,louizos2017causal}, and IBM (1,000, 177) \citep{shimoni2018benchmarking}. Each provides treatment assignments, factual outcomes, and response surfaces $\mu_0$ and $\mu_1$.

For multivariate evaluation, we construct multiple outcomes per individual while preserving covariates and treatment assignments. IHDP-Vector combines the first $d_Y\in\{2,5,10\}$ CEVAE/NPCI simulations for the same 747 individuals, while ACIC16-Vector similarly combines $d_Y$ simulated outcomes under a fixed assignment. MorphoMNIST-Intensity uses $28\times28$ images of the same digit under low- and high-intensity interventions, yielding 784-dimensional outcomes. Treatment depends on observed digit identity and stroke thickness, inducing confounding while preserving overlap. Models train only on factual outcomes, with both potential outcomes reserved for evaluation.

Paired counterfactual evaluation also requires shared noise across treatment arms, beyond marginal response surfaces. For $Y^{(A)}=f(X,A)+\epsilon$ with common $\epsilon$, we recover $\epsilon$ from the factual observation and compute $Y^{(1-A)}=f(X,1-A)+\epsilon$. For reproducibility, we fix the simulator versions and preprocessing for all benchmarks.

\subsection{Final Configurations and Evaluation Protocol}
\label{appendix:selection-protocol}

All results in Tables~\ref{tab:ipw-stable} and~\ref{tab:vector-results}
use five independent runs with random seeds 42--46. Each entry reports the
mean over these runs, and the parenthesized value is the sample standard
deviation across the same five runs.

For scalar outcomes, the final $(q,\alpha)$ pairs are $(.99,.25)$ for ACIC16,
$(.95,.75)$ for ACIC18, $(.95,.10)$ for IHDP, and $(.95,1.00)$ for IBM. For vector
outcomes, they are $(.95,.50)$, $(.90,.75)$, and $(.95,.50)$ for IHDP with
$d_Y=2,5,10$, respectively, and $(.95,.75)$ for all three ACIC16 dimensions. The
MorphoMNIST setting is $(.95,.50)$.

Outcome coordinates in the vector experiments are standardized using training factual
outcomes, and image pixels are scaled to $[-1,1]$. ACIC16-Vector and MorphoMNIST
standardize covariates using training-row means and standard deviations; IHDP-Vector uses
the original covariates without additional standardization. Each predicted conditional mean \(\widehat\mu_{i,a}\) is computed by averaging 100 PO-Flow samples under treatment \(a\) for scalar and vector outcomes, and 20 for MorphoMNIST. The fixed-step RK4 ODE solver uses 100 steps for vector outcomes and 50 for MorphoMNIST; the corresponding sliced-Wasserstein metrics use 200 and 100 random projection directions, respectively.

\paragraph{Regularized IPW weights.}
We regularize IPW weights before training to avoid unstable extreme weights. First, we clip
the cross-fitted propensity estimates:
\[
\widetilde e_i
=
\min\{1-\varepsilon,\max(\varepsilon,\widehat e_i)\},
\]
The corresponding inverse-probability weights are
\[
\widetilde r_i
=
\frac{A_i}{\widetilde e_i}
+
\frac{1-A_i}{1-\widetilde e_i}.
\]
Let $\widehat c_q$ be the empirical $q$-quantile of
$\{\widetilde r_i\}_{i=1}^n$ on the training set. We truncate large weights, then normalize
the result to have mean one:
\[
r_i^{(q)}
=
\min(\widetilde r_i,\widehat c_q),
\qquad
\bar r_i
=
\frac{r_i^{(q)}}
{n^{-1}\sum_{j=1}^n r_j^{(q)}}.
\]
Finally, we shrink the weights toward uniform weighting:
\[
w_i(\alpha)
=
1+\alpha(\bar r_i-1),
\qquad
0\leq\alpha\leq1.
\]
Thus \(\alpha=0\) gives uniform weights, whereas \(\alpha=1\) uses the truncated and normalized IPW weights without shrinkage. Clipping, truncation, and shrinkage limit the influence of extreme estimated weights, but they trade exact IPW correction for more stable finite-sample training. They cannot recover information where overlap is absent.

\subsection{Auxiliary Mean Readout}
\label{appendix:auxiliary-mean}

For ACIC16-Vector, the released implementation additionally trains an auxiliary mean readout $h_\theta(x,a)$ using factual squared error:
\[
\widehat{\mathcal L}_{\mathrm{joint}}(\theta)
=\widehat{\mathcal L}_{\mathrm{IPW}}(\theta)
+\frac{\lambda}{n}\sum_{i=1}^n
w_i\|h_\theta(x_i,a_i)-y_i\|_2^2,
\qquad \lambda=2.5.
\]
The readout reuses the neural network at the task input $(y,t)=(0,-1)$; it does not require counterfactual training labels. In this setting, mean predictions use $h_\theta$ rather than Monte Carlo flow averages. This auxiliary estimator need not equal the mean of the flow-generated distribution. The other final PO-Flow configurations use the flow-matching objective without this auxiliary term.

\subsection{Frozen Network Architectures and Hardware}
\label{appendix: nn structure and implementation}

We use the same velocity-network architecture for all scalar experiments. The network takes
$y_t$, $t$, $x$, and $a$ as inputs. It maps $y_t$ to $c=d_X+1$ features and uses an MLP to
encode $t$. A FiLM layer and two gated residual blocks use the time features and $[x,a]$ as
conditioning information. The skip outputs from the two blocks are summed, normalized by
$1/\sqrt 2$, projected to width 128, and passed to separate output heads for the two treatments.

For vector and image outcomes, a 32-dimensional feature map $\gamma(t)$ contains raw time
and Fourier features. With core width $w$, the network forms
\begin{equation}
c_\theta=\operatorname{MLP}_\theta([x,a,\gamma(t)]),
\qquad s_0=W_y y_t+c_\theta.
\end{equation}
Residual block $\ell$ applies conditional scale and shift followed by a width-$2w$ MLP:
\begin{equation}
\begin{aligned}
(g_\ell,b_\ell)&=W_{c,\ell}c_\theta,\\
u_\ell&=\operatorname{LN}(s_\ell)\odot
\{1+0.1\tanh(g_\ell)\}+b_\ell,\\
s_{\ell+1}&=\frac{s_\ell+W_{2,\ell}
\operatorname{SiLU}(W_{1,\ell}u_\ell)}{\sqrt 2}.
\end{aligned}
\end{equation}
Separate treatment heads map $\operatorname{SiLU}(\operatorname{LN}(s_L))$ to vector
velocities. The ACIC16-Vector mean readout reuses this same field at $(y,t)=(0,-1)$.

The final scalar networks use $h=128$ and $L=2$, with
$c=83,178,26,178$ for ACIC16, ACIC18, IHDP, and IBM, respectively. All tabular vector
networks use $w=192$: IHDP uses $L=4,3,3$ and ACIC16 uses $L=3,3,4$ for
$d_Y=2,5,10$, respectively. MorphoMNIST uses $w=512$ and $L=3$. Every vector and image
network uses a 32-dimensional time embedding, residual expansion factor two, and separate
treatment heads.

The journal runs used individual NVIDIA Tesla V100 accelerators with 
32 GB of memory. Our codebase records the complete configuration, software
environment, command, seed, and run time for each experiment. Propensity regressions are
fitted separately.

\subsection{Overlap Diagnostics}
\label{appendix:overlap-diagnostics}

This section checks whether regularized IPW improves covariate balance without
placing most of the weight on a small number of observations. We use two diagnostics of the
original overlap. AUC is the out-of-fold area under the ROC curve for predicting $A$ from
$X$, equivalently the probability that a treated observation receives a higher estimated
propensity than a control observation. An AUC of $.5$ means that the fitted model cannot
distinguish the groups, whereas a larger value indicates weaker overlap because treatment
is more predictable. The clipped fraction is the proportion of propensity estimates below
$.01$ or above $.99$; a large value indicates that many observations are close to violating
positivity.

We assess the final weights using
$\mathrm{ESS}/n=(\sum_i w_i)^2/(n\sum_i w_i^2)$ and $\max_i w_i$. Uniform weights give
$\mathrm{ESS}/n=1$, and concentration on fewer observations lowers this ratio. Because the
weights have mean one, $\max_i w_i$ is the largest individual weight relative to the
average. To measure balance for covariate $k$, we use
$\mathrm{SMD}_k=|\bar X_{1k}-\bar X_{0k}|/\sqrt{(s_{1k}^2+s_{0k}^2)/2}$. The tables report
$\max_k\mathrm{SMD}_k$; a decrease means that the largest observed mean imbalance has been
reduced. All entries are five-seed means.

\begin{table}[H]
\centering
\caption{Overlap diagnostics for the final scalar IPW settings.}
\label{tab:overlap-diagnostics}
\scriptsize
\setlength{\tabcolsep}{2.4pt}
\begin{tabular}{lcccccc}
\toprule
Data & $(q,\alpha)$ & AUC & clipped & ESS/$n$ & max $w$ & max SMD before$\to$after\\
\midrule
ACIC16 & $(.99,.25)$ & .6002 & .073\% & .9785 & 1.611 & .2490$\to$.1903\\
ACIC18 & $(.95,.75)$ & .5575 & .128\% & .9796 & 1.260 & .4517$\to$.3406\\
IHDP   & $(.95,.10)$ & .6659 & 3.585\% & .9937 & 1.255 & .4198$\to$.3980\\
IBM    & $(.95,1.0)$ & .9483 & 36.975\% & .7828 & 2.650 & 1.2824$\to$1.0483\\
\bottomrule
\end{tabular}
\end{table}

The scalar diagnostics show stable weights for ACIC16, ACIC18, and IHDP:
$\mathrm{ESS}/n$ is at least $.9785$, and the maximum SMD decreases in each case. IBM is
different. Its AUC is $.9483$, $36.975\%$ of propensity estimates are clipped, and the
maximum SMD remains above one after weighting. Thus treatment groups remain substantially
separated in IBM; increasing the weights would raise variance without restoring overlap.

\begin{table}[H]
\centering
\caption{IPW overlap diagnostics for the vector families. IHDP and ACIC16 share
$X,A$ across output dimensions, but IHDP uses dimension-specific weight strength.}
\label{tab:vector-overlap}
\scriptsize
\setlength{\tabcolsep}{2.8pt}
\begin{tabular}{lccccc}
\toprule
Data & $(q,\alpha)$ & AUC & ESS/$n$ & max $w$ & max SMD before$\to$after\\
\midrule
IHDP $d_Y=2$ & $(.95,.50)$ & .6991 & .8728 & 2.220 & .4384$\to$.3529\\
IHDP $d_Y=5$ & $(.90,.75)$ & .6991 & .8541 & 2.006 & .4384$\to$.3447\\
IHDP $d_Y=10$ & $(.95,.50)$ & .6991 & .8728 & 2.220 & .4384$\to$.3529\\
ACIC16 & $(.95,.75)$ & .7705 & .7653 & 2.713 & 1.1332$\to$.7526\\
MorphoMNIST & $(.95,.50)$ & .6663 & .9811 & 1.332 & .5055$\to$.2875\\
\bottomrule
\end{tabular}
\end{table}

For the vector and image settings, weighting reduces the maximum SMD in every
case. The remaining imbalance is largest for ACIC16, where the maximum SMD falls from
$1.1332$ to $.7526$ but remains substantial. The ESS ratios show that these reductions do
not arise from a few observations carrying nearly all the weight.

\subsection{Matched Baseline Adaptations}
\label{appendix:baseline-adaptations}

The reference baseline implementations use scalar TensorFlow-1, Keras, or Edward models.
Our PyTorch adaptations use the same data, splits, disclosed ground truth, and seeds as
PO-Flow; all standardization statistics are fitted on training rows only. For every neural
baseline, we adjust the hidden widths so that the number of trainable parameters matches
PO-Flow in the corresponding experiment while retaining the method's defining architecture.

TARNet and DragonNet use fixed settings within each outcome type: 120 epochs and learning
rate $10^{-3}$ with batch size 256 for scalar data and 512 for vector data; image runs use
30 epochs, batch size 512, and learning rate $5\times10^{-4}$. Both use
Adam with weight decay $10^{-4}$ and training-only preprocessing matched to PO-Flow.

TARNet uses two shared ELU layers and two outcome heads with two ELU layers each; CFR
matches this topology and adds a linear-MMD balance penalty. DragonNet uses three shared
ELU layers, two two-layer outcome heads, a treatment head, and targeted regularization with
one correction coefficient per outcome coordinate. GANITE uses vector generator,
discriminator, and inference networks. CEVAE uses a factorized-Gaussian vector likelihood
with latent generative and recognition networks. The artifact records the upstream commits
and every adaptation.

\subsection{Hutchinson Trace Estimator}
\label{appendix:Hutchinson}
Log-density evaluation requires the divergence of the velocity field. Let
$J_v(y)=\nabla_y v_\theta(y,t,x,a)$. If $\epsilon$ has zero mean and identity covariance,
Hutchinson's identity gives
$\operatorname{tr}J_v(y)=\mathbb E[\epsilon^\top J_v(y)\epsilon]$. We approximate the
Jacobian--vector product by finite differences
\citep{hutchinson1989stochastic,xu2023normalizing}:
\begin{equation}
\nabla \cdot v_{\theta}(y(t),t,x,a) \approx
\mathbb{E}_{p(\epsilon)}\!\left[
\epsilon^\top
\frac{v_{\theta}(y(t)+\sigma\epsilon,t,x,a)-v_{\theta}(y(t),t,x,a)}{\sigma}
\right],
\end{equation}
where $p(\epsilon)$ is a distribution on $\mathbb R^{d_Y}$ with
$\mathbb E[\epsilon]=0$ and $\operatorname{Cov}(\epsilon)=I$. The trace identity is exact;
the finite-difference approximation converges to it as $\sigma\to0$ under standard
smoothness conditions.

\section{Scalar Density Evaluation}
\label{appendix:scalar-density}

We evaluate scalar density estimation under the 10-fold protocol of~\citep{wu2026flow}.
For the semi-synthetic law $Y^{(a)}\sim\mathcal N(\mu_X(a),1)$, we use the likelihood in
Proposition~\ref{log density} and measure density accuracy by
\[
\mathrm{KL}\!\left(\hat{Y}^{(a)} \,\|\, Y^{(a)}\right)
=
\mathbb{E}_{\hat{y}\sim p_{\theta,Y\mid X,A}}
\left[
\log\frac{p_{\theta,Y\mid X,A}(\hat{y}\mid x,a)}{\mathcal{N}(\hat{y};\mu_X(a),1)}
\right],
\]
where $p_{\theta,Y\mid X,A}$ is evaluated by~\eqref{equation log density} using the trace
estimator in Appendix~\ref{appendix:Hutchinson}. Table~\ref{tab:kl_divergence} shows lower
KL divergence for PO-Flow than for INF~\citep{melnychuk2023normalizing} and
GP-ITE~\citep{alaa2017bayesian,huang2023gpmatch} on all four benchmarks.

\begin{table}[h!]
\centering
\caption{KL divergence of learned potential outcome densities
(10-fold cross-validation). Parentheses indicate standard deviation.}
\label{tab:kl_divergence}
\footnotesize
\setlength{\tabcolsep}{3pt}
\begin{tabular}{lcccc}
\toprule
\textbf{Method}
& \small\textbf{ACIC16}
& \small\textbf{ACIC18}
& \small\textbf{IHDP}
& \small\textbf{IBM} \\
\midrule

\textbf{PO-Flow}
& \textbf{0.14}{\tiny(0.04)}
& \textbf{0.08}{\tiny(0.02)}
& \textbf{0.09}{\tiny(0.03)}
& \textbf{0.08}{\tiny(0.03)} \\

INF
& 0.22{\tiny(0.09)}
& 0.15{\tiny(0.06)}
& 0.14{\tiny(0.07)}
& 0.10{\tiny(0.04)} \\

GP-ITE
& 1.73{\tiny(0.34)}
& 0.73{\tiny(0.19)}
& 1.42{\tiny(0.30)}
& 0.68{\tiny(0.17)} \\

\bottomrule
\end{tabular}
\end{table}

\bibliographystyle{IEEEtran}
\bibliography{references}

\begin{thebibliography}{10}
\providecommand{\url}[1]{#1}
\csname url@samestyle\endcsname
\providecommand{\newblock}{\relax}
\providecommand{\bibinfo}[2]{#2}
\providecommand{\BIBentrySTDinterwordspacing}{\spaceskip=0pt\relax}
\providecommand{\BIBentryALTinterwordstretchfactor}{4}
\providecommand{\BIBentryALTinterwordspacing}{\spaceskip=\fontdimen2\font plus
\BIBentryALTinterwordstretchfactor\fontdimen3\font minus
  \fontdimen4\font\relax}
\providecommand{\BIBforeignlanguage}[2]{{%
\expandafter\ifx\csname l@#1\endcsname\relax
\typeout{** WARNING: IEEEtran.bst: No hyphenation pattern has been}%
\typeout{** loaded for the language `#1'. Using the pattern for}%
\typeout{** the default language instead.}%
\else
\language=\csname l@#1\endcsname
\fi
#2}}
\providecommand{\BIBdecl}{\relax}
\BIBdecl

\bibitem{rubin1974estimating}
D.~B. Rubin, ``Estimating causal effects of treatments in randomized and
  nonrandomized studies,'' \emph{Journal of Educational Psychology}, vol.~66,
  no.~5, pp. 688--701, 1974.

\bibitem{holland1986statistics}
P.~W. Holland, ``Statistics and causal inference,'' \emph{Journal of the
  American statistical Association}, vol.~81, no. 396, pp. 945--960, 1986.

\bibitem{imbens2004nonparametric}
G.~W. Imbens, ``Nonparametric estimation of average treatment effects under
  exogeneity: A review,'' \emph{Review of Economics and statistics}, vol.~86,
  no.~1, pp. 4--29, 2004.

\bibitem{athey2016recursive}
S.~Athey and G.~Imbens, ``Recursive partitioning for heterogeneous causal
  effects,'' \emph{Proceedings of the National Academy of Sciences}, vol. 113,
  no.~27, pp. 7353--7360, 2016.

\bibitem{kunzel2019metalearners}
S.~R. K{\"u}nzel, J.~S. Sekhon, P.~J. Bickel, and B.~Yu, ``Metalearners for
  estimating heterogeneous treatment effects using machine learning,''
  \emph{Proceedings of the national academy of sciences}, vol. 116, no.~10, pp.
  4156--4165, 2019.

\bibitem{feuerriegel2024causal}
S.~Feuerriegel, D.~Frauen, V.~Melnychuk, J.~Schweisthal, K.~Hess, A.~Curth,
  S.~Bauer, N.~Kilbertus, I.~S. Kohane, and M.~van~der Schaar, ``Causal machine
  learning for predicting treatment outcomes,'' \emph{Nature Medicine},
  vol.~30, no.~4, pp. 958--968, 2024.

\bibitem{zhang2013assessing}
Z.~Zhang, C.~Wang, L.~Nie, and G.~Soon, ``Assessing the heterogeneity of
  treatment effects via potential outcomes of individual patients,''
  \emph{Journal of the Royal Statistical Society Series C: Applied Statistics},
  vol.~62, no.~5, pp. 687--704, 2013.

\bibitem{johansson2022generalization}
F.~D. Johansson, U.~Shalit, N.~Kallus, and D.~Sontag, ``Generalization bounds
  and representation learning for estimation of potential outcomes and causal
  effects,'' \emph{Journal of Machine Learning Research}, vol.~23, no. 166, pp.
  1--50, 2022.

\bibitem{johansson2016learning}
F.~Johansson, U.~Shalit, and D.~Sontag, ``Learning representations for
  counterfactual inference,'' in \emph{International conference on machine
  learning}.\hskip 1em plus 0.5em minus 0.4em\relax PMLR, 2016, pp. 3020--3029.

\bibitem{lei2021conformal}
L.~Lei and E.~J. Cand{\`e}s, ``Conformal inference of counterfactuals and
  individual treatment effects,'' \emph{Journal of the Royal Statistical
  Society Series B: Statistical Methodology}, vol.~83, no.~5, pp. 911--938,
  2021.

\bibitem{nasr2023counterfactual}
A.~Nasr-Esfahany, M.~Alizadeh, and D.~Shah, ``Counterfactual identifiability of
  bijective causal models,'' in \emph{International conference on machine
  learning}.\hskip 1em plus 0.5em minus 0.4em\relax PMLR, 2023, pp.
  25\,733--25\,754.

\bibitem{de2024transport}
L.~De~Lara, A.~Gonz{\'a}lez-Sanz, N.~Asher, L.~Risser, and J.-M. Loubes,
  ``Transport-based counterfactual models,'' \emph{Journal of Machine Learning
  Research}, vol.~25, no. 136, pp. 1--59, 2024.

\bibitem{charpentier2023optimal}
A.~Charpentier, E.~Flachaire, and E.~Gallic, ``Optimal transport for
  counterfactual estimation: A method for causal inference,'' in \emph{Optimal
  Transport Statistics for Economics and Related Topics}.\hskip 1em plus 0.5em
  minus 0.4em\relax Springer, 2024, pp. 45--89.

\bibitem{balakrishnan2025conservative}
S.~Balakrishnan, E.~H. Kennedy, and L.~Wasserman, ``Conservative inference for
  counterfactuals,'' \emph{Journal of Causal Inference}, vol.~13, no.~1, p.
  20230071, 2025.

\bibitem{chen2018neural}
R.~T. Chen, Y.~Rubanova, J.~Bettencourt, and D.~K. Duvenaud, ``Neural ordinary
  differential equations,'' \emph{Advances in neural information processing
  systems}, vol.~31, 2018.

\bibitem{lipman2023flow}
Y.~Lipman, R.~T.~Q. Chen, H.~Ben-Hamu, M.~Nickel, and M.~Le, ``Flow matching
  for generative modeling,'' in \emph{International Conference on Learning
  Representations (ICLR)}, 2023.

\bibitem{albergo2022building}
M.~S. Albergo and E.~Vanden-Eijnden, ``Building normalizing flows with
  stochastic interpolants,'' \emph{arXiv preprint arXiv:2209.15571}, 2022.

\bibitem{melnychuk2023normalizing}
V.~Melnychuk, D.~Frauen, and S.~Feuerriegel, ``Normalizing flows for
  interventional density estimation,'' in \emph{International Conference on
  Machine Learning}.\hskip 1em plus 0.5em minus 0.4em\relax PMLR, 2023, pp.
  24\,361--24\,397.

\bibitem{ma2024diffpo}
Y.~Ma, V.~Melnychuk, J.~Schweisthal, and S.~Feuerriegel, ``Diffpo: A causal
  diffusion model for learning distributions of potential outcomes,'' in
  \emph{Advances in Neural Information Processing Systems (NeurIPS)}, 2024.

\bibitem{wu2024counterfactual}
S.~Wu, W.~Zhou, M.~Chen, and S.~Zhu, ``Counterfactual generative models for
  time-varying treatments,'' in \emph{Proceedings of the 30th ACM SIGKDD
  Conference on Knowledge Discovery and Data Mining}, 2024, pp. 3402--3413.

\bibitem{wu2026flow}
D.~Wu, Y.~Xie, and D.~I. Inouye, ``Flow-based generative modeling of potential
  outcomes and counterfactuals,'' in \emph{2026 IEEE International Symposium on
  Information Theory (ISIT)}.\hskip 1em plus 0.5em minus 0.4em\relax IEEE,
  2026, pp. 1--6.

\bibitem{curth2021nonparametric}
A.~Curth and M.~Van~der Schaar, ``Nonparametric estimation of heterogeneous
  treatment effects: From theory to learning algorithms,'' in
  \emph{International Conference on Artificial Intelligence and
  Statistics}.\hskip 1em plus 0.5em minus 0.4em\relax PMLR, 2021, pp.
  1810--1818.

\bibitem{kennedy2023towards}
E.~H. Kennedy, ``Towards optimal doubly robust estimation of heterogeneous
  causal effects,'' \emph{Electronic Journal of Statistics}, vol.~17, no.~2,
  pp. 3008--3049, 2023.

\bibitem{hahn2020bayesian}
P.~R. Hahn, J.~S. Murray, and C.~M. Carvalho, ``Bayesian regression tree models
  for causal inference: Regularization, confounding, and heterogeneous effects
  (with discussion),'' \emph{Bayesian Analysis}, vol.~15, no.~3, pp. 965--1056,
  2020.

\bibitem{shalit2017estimating}
U.~Shalit, F.~D. Johansson, and D.~Sontag, ``Estimating individual treatment
  effect: generalization bounds and algorithms,'' in \emph{International
  conference on machine learning}.\hskip 1em plus 0.5em minus 0.4em\relax PMLR,
  2017, pp. 3076--3085.

\bibitem{shi2019adapting}
C.~Shi, D.~M. Blei, and V.~Veitch, ``Adapting neural networks for the
  estimation of treatment effects,'' in \emph{Advances in Neural Information
  Processing Systems}, vol.~32, 2019.

\bibitem{yoon2018ganite}
J.~Yoon, J.~Jordon, and M.~Van Der~Schaar, ``Ganite: Estimation of
  individualized treatment effects using generative adversarial nets,'' in
  \emph{International conference on learning representations}, 2018.

\bibitem{louizos2017causal}
C.~Louizos, U.~Shalit, J.~M. Mooij, D.~Sontag, R.~Zemel, and M.~Welling,
  ``Causal effect inference with deep latent-variable models,'' \emph{Advances
  in neural information processing systems}, vol.~30, 2017.

\bibitem{bica2020estimating}
I.~Bica, A.~M. Alaa, J.~Jordon, and M.~van~der Schaar, ``Estimating
  counterfactual treatment outcomes over time through adversarially balanced
  representations,'' \emph{arXiv preprint arXiv:2002.04083}, 2020.

\bibitem{melnychuk2022causal}
V.~Melnychuk, D.~Frauen, and S.~Feuerriegel, ``Causal transformer for
  estimating counterfactual outcomes,'' in \emph{International conference on
  machine learning}.\hskip 1em plus 0.5em minus 0.4em\relax PMLR, 2022, pp.
  15\,293--15\,329.

\bibitem{pawlowski2020deep}
N.~Pawlowski, D.~Coelho~de Castro, and B.~Glocker, ``Deep structural causal
  models for tractable counterfactual inference,'' \emph{Advances in neural
  information processing systems}, vol.~33, pp. 857--869, 2020.

\bibitem{javaloy2023causal}
A.~Javaloy, P.~S{\'a}nchez-Mart{\'\i}n, and I.~Valera, ``Causal normalizing
  flows: from theory to practice,'' \emph{Advances in Neural Information
  Processing Systems}, vol.~36, pp. 58\,833--58\,864, 2023.

\bibitem{xia2023neural}
K.~Xia, Y.~Pan, and E.~Bareinboim, ``Neural causal models for counterfactual
  identification and estimation,'' in \emph{International Conference on
  Learning Representations}, 2023.

\bibitem{sanchez2022diffusion}
P.~Sanchez and S.~A. Tsaftaris, ``Diffusion causal models for counterfactual
  estimation,'' \emph{arXiv preprint arXiv:2202.10166}, 2022.

\bibitem{ribeiro2025dynamic}
F.~De~Sousa~Ribeiro, A.~Santhirasekaram, and B.~Glocker, ``Counterfactual
  identifiability via dynamic optimal transport,'' in \emph{Advances in Neural
  Information Processing Systems}, vol.~38, 2025, pp. 164\,117--164\,165.

\bibitem{hirano2003efficient}
K.~Hirano, G.~W. Imbens, and G.~Ridder, ``Efficient estimation of average
  treatment effects using the estimated propensity score,''
  \emph{Econometrica}, vol.~71, no.~4, pp. 1161--1189, 2003.

\bibitem{cole2008constructing}
S.~R. Cole and M.~A. Hern{\'a}n, ``Constructing inverse probability weights for
  marginal structural models,'' \emph{American Journal of Epidemiology}, vol.
  168, no.~6, pp. 656--664, 2008.

\bibitem{zhang2012identifiability}
K.~Zhang and A.~Hyvarinen, ``On the identifiability of the post-nonlinear
  causal model,'' \emph{arXiv preprint arXiv:1205.2599}, 2012.

\bibitem{strobl2023identifying}
E.~V. Strobl and T.~A. Lasko, ``Identifying patient-specific root causes with
  the heteroscedastic noise model,'' \emph{Journal of Computational Science},
  vol.~72, p. 102099, 2023.

\bibitem{rosenblatt1952remarks}
M.~Rosenblatt, ``Remarks on a multivariate transformation,'' \emph{The annals
  of mathematical statistics}, vol.~23, no.~3, pp. 470--472, 1952.

\bibitem{knothe1957contributions}
H.~Knothe, ``Contributions to the theory of convex bodies.'' \emph{Michigan
  Mathematical Journal}, vol.~4, no.~1, pp. 39--52, 1957.

\bibitem{brenier1991polar}
Y.~Brenier, ``Polar factorization and monotone rearrangement of vector-valued
  functions,'' \emph{Communications on Pure and Applied Mathematics}, vol.~44,
  no.~4, pp. 375--417, 1991.

\bibitem{villani2009optimal}
C.~Villani, \emph{Optimal Transport: Old and New}.\hskip 1em plus 0.5em minus
  0.4em\relax Springer, 2009.

\bibitem{amos2017input}
B.~Amos, L.~Xu, and J.~Z. Kolter, ``Input convex neural networks,'' in
  \emph{International conference on machine learning}.\hskip 1em plus 0.5em
  minus 0.4em\relax PMLR, 2017, pp. 146--155.

\bibitem{makkuva2020optimal}
A.~Makkuva, A.~Taghvaei, S.~Oh, and J.~Lee, ``Optimal transport mapping via
  input convex neural networks,'' in \emph{International Conference on Machine
  Learning}.\hskip 1em plus 0.5em minus 0.4em\relax PMLR, 2020, pp. 6672--6681.

\bibitem{castro2019morphomnist}
D.~C. Castro, J.~Tan, B.~Kainz, E.~Konukoglu, and B.~Glocker, ``Morpho-mnist:
  quantitative assessment and diagnostics for representation learning,''
  \emph{Journal of Machine Learning Research}, vol.~20, no. 178, pp. 1--29,
  2019.

\bibitem{melistas2024benchmarking}
T.~Melistas, N.~Spyrou, N.~Gkouti, P.~Sanchez, A.~Vlontzos, Y.~Panagakis,
  G.~Papanastasiou, and S.~A. Tsaftaris, ``Benchmarking counterfactual image
  generation,'' in \emph{Advances in Neural Information Processing Systems:
  Datasets and Benchmarks Track}, 2024.

\bibitem{hill2011bayesian}
J.~L. Hill, ``Bayesian nonparametric modeling for causal inference,''
  \emph{Journal of Computational and Graphical Statistics}, vol.~20, no.~1, pp.
  217--240, 2011.

\bibitem{shimoni2018benchmarking}
Y.~Shimoni, C.~Yanover, E.~Karavani, and Y.~Goldschmnidt, ``Benchmarking
  framework for performance-evaluation of causal inference analysis,''
  \emph{arXiv preprint arXiv:1802.05046}, 2018.

\bibitem{hutchinson1989stochastic}
M.~F. Hutchinson, ``A stochastic estimator of the trace of the influence matrix
  for laplacian smoothing splines,'' \emph{Communications in
  Statistics-Simulation and Computation}, vol.~18, no.~3, pp. 1059--1076, 1989.

\bibitem{xu2023normalizing}
C.~Xu, X.~Cheng, and Y.~Xie, ``Normalizing flow neural networks by jko
  scheme,'' \emph{Advances in Neural Information Processing Systems}, vol.~36,
  pp. 47\,379--47\,405, 2023.

\bibitem{alaa2017bayesian}
A.~M. Alaa and M.~Van Der~Schaar, ``Bayesian inference of individualized
  treatment effects using multi-task gaussian processes,'' \emph{Advances in
  neural information processing systems}, vol.~30, 2017.

\bibitem{huang2023gpmatch}
B.~Huang, C.~Chen, J.~Liu, and S.~Sivaganisan, ``Gpmatch: A bayesian causal
  inference approach using gaussian process covariance function as a matching
  tool,'' \emph{Frontiers in Applied Mathematics and Statistics}, vol.~9, p.
  1122114, 2023.

\end{thebibliography}

\end{document}